\documentclass[1p]{elsarticle}

\let\today\relax
\makeatletter
\def\ps@pprintTitle{%
    \let\@oddhead\@empty
    \let\@evenhead\@empty
    \def\@oddfoot{\footnotesize\itshape
         {Accepted by Neurocomputing} \hfill\today}%
    \let\@evenfoot\@oddfoot
    }
\makeatother

\usepackage{tikz}
\usetikzlibrary{shapes,arrows,shadows,positioning}

\graphicspath{ {./figures/} }

\usepackage{mathtools}
\usepackage{amsthm}
\usepackage{amssymb}
\usepackage{siunitx} 
\usepackage{amsmath,bm,times}
\DeclareMathOperator*{\argmax}{argmax}
\DeclareMathOperator*{\argmin}{argmin}
\DeclareMathOperator{\trans}{{\mathrm{T}}}

\newcommand*\abs[1]{\left \lvert#1 \right \rvert}

\usepackage{breqn}

\usepackage{algorithm, algorithmic}

\usepackage{marginnote}
\usepackage{multirow,makecell}
\usepackage{tabularx}
\newcolumntype{Y}{>{\centering\arraybackslash}X}

\usepackage{array}
\usepackage{booktabs}
\usepackage{adjustbox}

\usepackage[figuresright]{rotating}
\usepackage{xcolor}

\newcommand*\revise[1]{\textcolor{black}{#1}}
\newcommand*\reviseS[1]{\textcolor{black}{#1}}

\usepackage{changes} 
\definechangesauthor[name={KunHe}, color=red]{He}
\definechangesauthor[name={Xinze}, color=red]{Zh}
\definechangesauthor[name={Bao}, color=blue]{bao}

\usepackage{color}

\usepackage{hyperref}

\usepackage[nameinlink,capitalise]{cleveref}
\Crefname{figure}{Figure }{Figures }

\usepackage{caption}
\usepackage{subcaption}

\newcommand\s{$^\star$}

\usepackage{pifont}
\newcommand{\cmark}{\ding{51}}
\newcommand{\xmark}{\ding{55}}

\biboptions{numbers,sort&compress}

\begin{document}

\begin{frontmatter}

    \title{Error-feedback Stochastic Modeling Strategy \\
         for Time Series Forecasting with Convolutional Neural Networks}

    \author[mainaddress]{Xinze Zhang\fnref{fn1}}

    \author[secondaryaddress]{Kun He\fnref{fn1}}

    \author[mainaddress]{Yukun Bao\corref{mycorrespondingauthor}}
    \cortext[mycorrespondingauthor]{Corresponding author}
    \ead{yukunbao@hust.edu.cn}

    \address[mainaddress]{Center for Modern Information Management, School of Management,\\Huazhong University of Science and Technology, Wuhan, 430074 CN}
    \address[secondaryaddress]{School of Computer Science \& Technology,\\Huazhong University of Science and Technology, Wuhan, 430074 CN}

    \fntext[fn1]{The first two authors contribute equally to this work.}

    \begin{abstract}
        Despite the superiority of convolutional neural networks demonstrated in time series modeling and forecasting, it has not been fully explored on the design of the neural network architecture and the tuning of the hyper-parameters. 
        Inspired by the incremental construction strategy for building a random multilayer perceptron, we propose a novel Error-feedback Stochastic Modeling (ESM) strategy to construct a random Convolutional Neural Network (ESM-CNN) for time series forecasting task, which builds the network architecture adaptively. 
        The ESM strategy suggests that random filters and neurons of the error-feedback fully connected layer are incrementally added to steadily compensate the prediction error during the construction process, and then a filter selection strategy is introduced to enable ESM-CNN to extract the different size of temporal features, providing helpful information at each iterative process for the prediction.
        The performance of ESM-CNN is justified on its prediction accuracy of one-step-ahead and multi-step-ahead forecasting tasks respectively.
        Comprehensive experiments on both the synthetic and real-world datasets show that the proposed ESM-CNN not only outperforms the state-of-art random neural networks, but also exhibits stronger predictive power and less computing overhead in comparison to trained state-of-art deep neural network models.
    \end{abstract}

    \begin{keyword}
        Convolutional neural network, error-feedback stochastic modeling, time series forecasting.
    \end{keyword}

\end{frontmatter}


\section{Introduction}~\label{sec:intro}
\revise{Recent studies have revealed that convolutional neural network (CNN), which benefits from its strength in extracting local features via multiple convolutional filters and learning representation by fully connected layers, has been successfully implemented for time series forecasting that is of great importance in real world applications, such as finance~\cite{sezer2018algorithmic,cavalliCNNbased2021}, energy~\cite{luo2019can,dongNovel2020}, and electric load~\cite{sadaei2019short,kuoHigh2018}.
}

{\revise{
The importance of time series forecasting and the applications of CNNs for modeling it have raised increasing attention to construct CNNs for time series prediction.
As for the model selection, constructing a CNN commonly follows a typical paradigm, which}
is to determine and fix the hyper-parameters (i.e., neural architecture, learning rate, and training epochs) first, and then train the network based on some gradient descent optimization methods, making it inflexible and extremely laborious to evolve the \revise{hyper-parameters to build CNNs
\cite{zela2018towards}.
Hence, constructing a powerful CNN for time series prediction efficiently is focused in this study.}}

\revise{
Reasearchers have investigated many methods on the model selection for building the forecasting model.
Hu et al.~\cite{huShortterm2016} presented a hybrid PSO-SVR forecasting method, which used particle swarm optimization (PSO) to search optimal support vector regression (SVR) parameters.
Flores et al.~\cite{floresEvolutive2012} evolved auto-regression moving average (ARMA) and multilayer perceptron (MLP) for time series forecasting by using genetic algorithm to indicate which variables are active in ARMA or to optimize the hidden neurons, activation functions and other configurations of MLP.
Further, Elsken et al.~\cite{elskenNeural2019a} provided a survey on searching the hyper-parameters of neural network (NN) models, which concludes the common practice of evolving NN is to set up a search space of hyper-parameters defining the network architecture and then search the optimal parameters.
}

\revise{
However, there exist imperfections in existing researches.
For example, during the search phase of evolving CNN models, the candidate model configurations (e.g., the number of the filters, the type of activation functions and pooling layers, the size of convolutional and pooling filters) are fixed and then the weight parameters of the defined architectures are trained with a number of epochs, only after that, the candidate hyper-parameters can be evaluated. Unfortunately, it is extremely time-consuming of training many different model configurations~\cite{baker2017accelerating}.}

In contrast, random CNN with untrained stochastic filters can be considered as an alternative option.
It is suggested that stochastic filters that are iteratively generated to one convolutional layer after another can perform as well as trained filters for image representation inverting, texture synthesis, and style transfer~\cite{he2016powerful}.
\revise{The authors built a CNN architecture called VGG~\footnote{{https://www.robots.ox.ac.uk/~vgg/research/very\_deep/}} with a stacked random strategy.} 
For each layer they sampled several sets of weights in Gaussian distribution, selected one set of weights with the lowest inverting loss, and fixed the weights of each layer in forwarding order.
Another try in audio identification showed that audio texture produced by trained one-dimensional CNN \revise{was} found inferior to those produced by a random CNN~\cite{antognini2019audio}.
Yu et al.\cite{yu2019impact} further showed that the precision of a CNN with random filters is close to a CNN of the same architecture but with trained filters in three scenarios of time series forecasting.
Despite that random CNN is comparable to pre-trained CNN in some deep learning tasks, however, it is still hard to model the time series steadily \revise{due to the inner stochasticity}.

In this study, \revise{w}e propose a novel error-feedback stochastic modeling based convolutional neural network (ESM-CNN) by incrementally generating new convolution filters and the corresponding fully connected layer's neuron to forecast the time series.
Different from the existing method that updates all the output weights~\cite{igelnik1995stochastic, wang2017stochastic}, in order to avoid overfitting, we iteratively add neurons of the error-feedback fully connected layer to the filters and calculate the parameters.
Furthermore, we propose a greedy based filter selection method to generate convolution filters with multiple filter sizes in constructing ESM-CNN, enabling ESM-CNN to extract different temporal patterns for the prediction.

The main contributions of this study are summarized as follows:
\begin{itemize}
      \item {
            We propose a novel error-feedback stochastic modeling strategy to \revise{efficiently craft a powerful} CNN, called ESM-CNN, for time series forecasting task. 
            \revise{
            An ESM-CNN is built by incrementally generating new random filters to promote its construction efficiency, and individually configuring the corresponding neurons of the error-feedback fully connected layer to theoretically guarantee its forecasting performance as well as improving its stability.
            }
            }
      \item {
            A greedy based filter selection method is introduced to select the convolution filters from the randomly generated candidates with different filter sizes, enabling ESM-CNN to extract different temporal patterns and \revise{select the convolution filters adaptively}.
            }
      \item {
            \revise{Comprehensive experiments} on simulated and real-world datasets demonstrate \revise{the efficiency and the effectiveness of the proposed ESC strategy as well as filter selection method} of ESM-CNN \revise{for} time series \revise{prediction}. ESM-CNN \revise{exhibits strong predictive power, with low computation overhead and good generalization ability} compared with standard random neural networks and deep neural networks.
            }
\end{itemize}

The rest of the paper is organized as follows.
Section 2 introduces related works on CNN and random MLP, indicating the motivation of our work.
Details about the proposed ESM-CNN modeling strategy are depicted in Section 3.
Section 4 presents details on dataset description, counterparts selection, accuracy measure, and experimental procedure.
Experimental results are presented and discussed in Section 5.
Finally, Section 6 concludes this work.

\section{Related works \label{sec:related}}
To clearly illustrate the motivation of the proposed error-feedback stochastic modeling strategy of constructing a CNN for time series prediction, the \reviseS{preliminary formulation and related works of implementing} CNN as well as random MLP for time series forecasting are introduced briefly.

\reviseS{
\subsection{Forecasting formulation}
The essence of time series forecasting is a type of function approximation procedure.
As defined in \cite{hewamalageRecurrent2021}, an univariate time series prediction problem refers to forecasting future values of a time series based on its past values.
That is, given a time series with $T$ observations $\{x_1,\dots, x_T\}$ and the next $H$-steps-ahead observations $\{x_{T+1},\ldots,x_{T+H}\}$, the univariate time series prediction problem can be formulated as learning a approximation function:
\begin{equation}
    f: f(x_1,\dots, x_T) + e = \{x_{T+1},\ldots,x_{T+H}\}
\end{equation}
to minimize the prediction error  $\| e \|$.
Here, $f$ is the function approximated by the model constructed for the time series forecasting with the variable $x$. 
The model forecasts the future values from $T+1$ to $T+H$, where $H$ is the forecasting horizon.
In the following, we give the formulation of implementing CNN and random MLP for time series forecasting.
}

\subsection{Convolutional Neural Network}
Convolutional neural network is developed by connecting filters to local fields on the input to perceptron~\cite{lecun1995convolutional}.
With local connections, filters can extract elementary features that are likely to be useful across the entire time series.
The outputs of such a set of filters constitute the feature maps~\cite{bengio2013representation}.
And then, these feature maps are combined in higher layers, i.e., fully connected layers, to learn abstract representations.
Here we briefly introduce a typical convolutional neural network for time series forecasting, which consists of a single convolutional layer to extract feature maps, a single pooling layer to subsample the feature maps, and a fully connected layer to learn the outputs.

\reviseS{Consider that a time series data $D=\left\{\left(X_{i}, Y_{i}\right) \in\left(\mathbb{R}^{T} \times \mathbb{R}^{H}\right)\right\}_{i=1}^{N}$ is composed of $N$ samples with $T$ observations $X_i = [x_1,\dots, x_T]^{\trans}$ for forecasting the next $H$-steps-ahead observations  $Y_i = [x_{T+1},\dots, x_{T+H}]^{\trans}$.}
The CNN with $C$ convolution filters can be expressed as\reviseS{:}
\begin{align}
    f_C &= \sum^C_{j=1} \left( \sum^{T-K+2}_{i=1}\beta_{j}^i p_j^i\right) + \beta_0,& &   \\
    p_j^i &= \frac{\sum^{K_p}_{k_p=1} m_j^{i+k_p-1}}{K_p}, \quad &i &= 1,..., T-K_p - K_m+2, \label{eq:pool} \\
    m_j^t &= \sigma \left(\sum^{K_m}_{k_m=1} w^{k_m}_j x_{t+k_m-1} +b_j\right), \quad &t &= 1,\dots, T-K_m +1,
\end{align}
where $K = K_p + K_m$,
$K_p$ and $K_m$ are the kernel size of the pooling and convolution operation, respectively.
$p_j$ denotes the feature map vector $[p_j^1, \ldots, p^{T-{K}+2}_j]^{\mathrm{T}}$ sequentially down-sampled from $m_j$ by \revise{the} average pooling,
${m}_j$ denotes the feature map vector $[m_j^1, \ldots, m^{T-{K_m}+1}_j]^{\mathrm{T}}$ sequentially extracted by the $j$-th filter with $K_m$ size of local connections from the input time series.
And $\sigma(\cdot)$ is a nonlinear activation function, set as sigmoid in this work,
${w_j} = [w_j^1,\ldots, w_j^{K_m}]$ and $b_j$ are the weight and bias of the $j$-th filter.

CNN usually uses the pooling operation to subsample the feature maps.
The advantage of this operation is to reduce the convolutional output band, and be more robust to variations in feature maps~\cite{zhao2017convolutional}.
Average or maximum pooling is \revise{commonly} utilized for time series analysis~\cite{zheng2014time,yang2015deep,koprinska2018convolutional}, and \revise{average pooling is selected in this study}.
Besides, $[\beta_1^1, \beta_1^2,\ldots,\beta_C^{T-K+2}]$ ($\beta_j^i = [\beta_{j}^{i,1},\ldots,\beta_{j}^{i,H}]^{\mathrm{T}}$) and $\beta_0$ are respectively the weights and bias in the fully connected layer that are linked to all feature map vectors.

Since all the weights and biases are gradient-based trained with back-propagation, CNN can extract features via filters from the input time series and linearly link these features to predictions by the fully connected layer.
However, the fixed architecture of trained CNN requires predefined hyper-parameters, which is computationally expensive to tune the number of filters, training epochs, and other \revise{configurations}.

\subsection{Random Multilayer Perceptron}

\begin{figure}[!ht]
    \begin{subfigure}[b]{0.25\textwidth}
        \includegraphics[height=\linewidth]{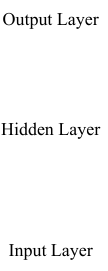}
        \caption*{}
    \end{subfigure}
    \hspace*{-0.15\textwidth}
    \begin{subfigure}[b]{0.25\textwidth}
        \includegraphics[height=\linewidth]{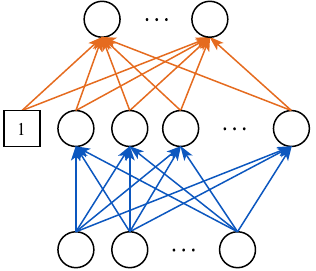}
        \caption{Schimidt's method}
    \end{subfigure}
    \hspace*{0.03\textwidth}
    \begin{subfigure}[b]{0.25\textwidth}
        \includegraphics[height=\linewidth]{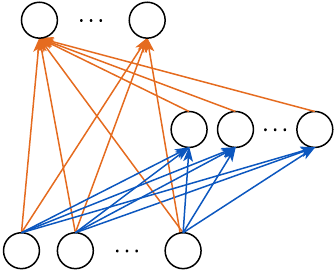}
        \caption{RVFL network}
    \end{subfigure}
    \hspace*{\fill}
    \begin{subfigure}[b]{0.25\textwidth}
        \includegraphics[height=\linewidth]{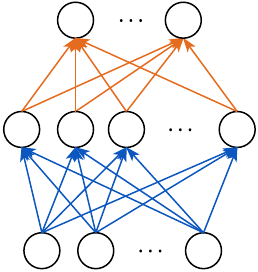}
        \caption{ELM network}
    \end{subfigure}
    \caption{\label{fig:randomnet}Typical random multilayer perceptron, where the blue arrows denote the randomly initialized and fixed hidden weights, and the yellow arrows represent the analytically determined output weights.}
\end{figure}

Random multilayer perceptron, which is initially proposed by~\cite{schmidt1992feed} and further developed on the random vector version of the functional-link (RVFL) network~\cite{igelnik1995stochastic} as well as extreme learning machine (ELM) network~\cite{huangExtreme2006}, uses the randomly initialized and fixed hidden layers and analytically computed output layer to build a MLP. The structure\revise{s} of these three typical random MLP\revise{s} \revise{are} shown in ~\autoref{fig:randomnet}.

Inspired by the pioneering work~\cite{schmidt1992feed,igelnik1995stochastic,huangExtreme2006}, many researches \revise{have been }\reviseS{conducted} for developing random MLPs.
\citet{paoLearning1994} indicated that a single step pseudo-inverse solution was sufficed for RVFL.
\citet{scardapaneDistributed2015} proposed the distributed learning algorithms for RVFL by leveraging the subsets of the data to train the local models and find the common output weights of the master model.
\citet{zhangUnsupervised2019} used a sparse and regular autoencoder to make an unsupervised learning RVFL network.
\citet{huangExtreme2012} investigated the relationship between ELM and support vector machine and extend ELM with kernel methods.
\citet{kasun2013representational} introduced orthogonal random feature mapping to build a ELM-based autoencoder, providing a promising solution to representation learning.

Since the conventional random MLPs are proposed and constructed in the deterministic method, it is also quite challenging to select the proper parameters, such as the number of the hidden units, for the practical implementations.
To resolve this issue, an incrementally constructive method has been leveraged to build random MLPs.
\citet{huang2006universal} proposed an incremental ELM (IELM) by iteratively adding the randomly generated neural unit to the hidden layer of the network and proved it can converge to any continuous target function.
\citet{wang2017stochastic} developed an incremental random MLP with stochastic configuration algorithms and named the model as stochastic configuration network (SCN), where the hidden layer was built incrementally with selected random neurons and the output layer was determined via the least squares method.

The implementation of incremental random MLP for time series modeling and forecasting is described below.
Assume a random MLP has generated $L$ hidden neurons in its single fully connected hidden layer\reviseS{:}
\begin{equation}
    f_{L} =\sum_{j=1}^{L} \beta_{j} g_{j}\left( \sum^T_{i=1}w_{j}^i x_i+b_j\right),
\end{equation}
where ${w_j} = [w_j^1,\ldots, w_j^T]$ and $b_j$ are respectively the weight and bias of the $j$-th neuron, $g(\cdot)$ is the sigmoid activation function.
And the neurons are represented as $[g_1,\ldots,g_L]^{\trans}$ with the corresponding output weights $[\beta_1, \ldots, \beta_L], \beta_{j}=\left[{\beta}_{j}^1, \ldots, \beta_{j}^H\right]^{\mathrm{T}}$, $H$ is the prediction horizons.
The prediction error is denoted as\reviseS{:}
\begin{equation}
    e_{L} = Y-f_{L} = [e_{L}^1, \ldots, e_{L}^H]^{\mathrm{T}}.
\end{equation}

If $\left\|e_{L}\right\|$ is higher than a tolerance level $\epsilon$, the random MLP incrementally add\reviseS{s} a new random neuron $g_{L+1}$ ($w_{L+1}$ and $b_{L+1}$) to the hidden layer and update\reviseS{s} the parameters of the output layer, where IELM directly generates a random hidden neuron and computes its corresponding output weight. In contrast to IELM, SCN generates a new selected random hidden neuron that satisfies a supervisory condition and update\revise{s} the whole parameters of the output layer.

So far, tremendous efforts have been devoted to develop theories and applications for random MLPs, and yield considerable performance for the regressions and predictions~\cite{huang2006universal,vanheeswijkAdaptive2009,huangExtreme2012,wang2017stochastic,zengSwitching2017,wang2018crude}.
However, there are few works related to random CNN in the literature of time series forecasting.
In this study, \revise{to craft a powerful CNN for time series prediction efficiently}, an error-feedback stochastic modeling strategy is proposed and justified within CNN.

\section{The ESM-CNN modeling strategy \label{sec:methodology}}
To address the limitation of implementing CNN for time series forecasting, 
we propose an error-feedback stochastic modeling strategy of building a random convolutional neural network, called ESM-CNN, for time series forecasting.
This section presents the implementation details and convergence analysis of ESM-CNN.

\subsection{Error-feedback Stochastic Strategy}
~\label{subsec:sc}
\revise{In order to efficiently construct a powerful CNN forecasting model, which benefits from the efficiency of random NNs and also tackles the instability of its inner randomness, we propose an error-feedback stochastic modeling strategy for constructing a random CNN.
An ESM-CNN is built by incrementally generating new random filters to promote its efficiency, and individually configuring the corresponding neurons of the error-feedback fully connected layer to guarantee its forecasting performance with theoretical convergence property.}

Specifically, assume that an ESM-CNN with $C$ filters has been constructed within a single convolutional layer, which can be expressed as\reviseS{:}
\begin{equation}
    \label{eq:sccnn}
    f_C = \sum^C_{j=1} \left( \sum^{T-K+2}_{i=1}\beta_{j}^i p_j^i + \beta^0_j \right),
\end{equation}
and briefly written as $f_C= \sum^C_{j=1}\sum^{T-K+2}_{i=0} \beta_j^i p_j^i$, where $\beta_j = [\beta_j^0, \beta_j^1, \ldots, \beta_j^{T-K+2}]$ present the weight parameter and bias parameter of the fully connected layer linked with the $j$th subsampled filter, $p_j^0 = I$.
Let the prediction error of ESM-CNN be denoted as
\begin{equation}
    e_C = Y- f_C = [e_C^1,\ldots, e_C^H].
\end{equation}

If the mean square error has not reached the tolerance level $\epsilon$, ESM will generate a stochastic filter $m_{C+1} $ with the corresponding pooling values $p_{C+1}$ and fully connected layer neurons to CNN, where the parameters of the newly added fully connected layer neurons are individually calculated under the error-feedback from $e_C$ via the least squares method\reviseS{:}
\begin{equation}\label{eq:scupdate}
    \left[\beta_{{C+1}}^0, \ldots, \beta_{{C+1}}^{T-K+2} \right]=\argmin _{\beta}\|e_C -\sum_{i=0}^{T-K+2} \beta_{C+1}^i p_{C+1}^i \|.
\end{equation}
Through this strategy, it is guaranteed that the prediction error of the CNN is monotonically decreasing and converged, 
\reviseS{which is proved as follows. And the detailed manipulations for the derived formulas are provided in \ref{app:proof}.}
\begin{proof}
    The intermediate prediction error sequence $\tilde{e}_{C+1}^{\, 0}, \ldots, \tilde{e}_{C+1}^{\, T-K+2} $ and the intermediate parameters $\tilde{\beta}_{C+1}^{\, 0}, \ldots, \tilde{\beta}_{C+1}^{\, T-K+2}$ of the new added fully connected layer  are introduced as\reviseS{:}
    \revise{
        \begin{alignat*}{2}
         & \tilde{e}_{C+1}^{\, i+1}       & = & \mspace{18mu} \tilde{e}_{C+1}^{\, i}-\tilde{\beta}_{C+1}^{\, i+1} p_{C+1}^{i+1}, \quad i = 0,\ldots,T-K+1,                                        \\
        \shortintertext{where}
         & \tilde{\beta}_{C+1}^{\, i+1}   & = & \mspace{18mu} [\tilde{\beta}_{C+1}^{\, i+1,1}, \ldots,\tilde{\beta}_{C+1}^{\, i+1,h},\ldots, \tilde{\beta}_{C+1}^{\, i+1,H}],                     \\
         & \tilde{\beta}_{C+1}^{\, i+1,h} & = & \mspace{18mu} \left\langle \tilde{e}_{C+1}^{\, i,h}, p_{C+1}^{\, i+1}\right\rangle /\left\|p_{C+1}^{\, i+1}\right\|^{2} , \quad h= 1, \ldots, H , \\
        \shortintertext{and}
         & \tilde{e}_{C+1}^{\, 0}         & = & \mspace{18mu} e_{C}- \tilde{\beta}_{C+1}^{\, 0} p_{C+1}^{\, 0},                                                                                   \\
         & \tilde{\beta}_{C+1}^{\, 0}   & = & \mspace{18mu} [\tilde{\beta}_{C+1}^{\, 0,1}, \ldots,\tilde{\beta}_{C+1}^{\, 0,h},\ldots, \tilde{\beta}_{C+1}^{\, 0,H}],                     \\
         & \tilde{\beta}_{C+1}^{\, 0,h} & = & \mspace{18mu} \left\langle {e}_{C}^{\, h}, p_{C+1}^{\, 0}\right\rangle /\left\|p_{C+1}^{\, 0}\right\|^{2} , \quad h= 1, \ldots, H , \\
         \shortintertext{making}
         & \tilde{e}_{C+1}^{\, T-K+2} \,  & = & \mspace{18mu} {e}_{C} - \sum^{T-K+2}_{i=0} \tilde{\beta}_{C+1}^{\, i} p_{C+1}^{\, i}.
    \end{alignat*}
    }
    \revise{Since the parameters of the new added error-feedback fully connected layer are calculated by the least square method:
    $$
        \left[\beta_{{C+1}}^0, \ldots, \beta_{{C+1}}^{T-K+2} \right]=\argmin _{\beta} \,\|e_C -\sum_{i=0}^{T-K+2} \beta_{C+1}^i p_{C+1}^i \|,
    $$
    the basic inequality between $\left\|e_{C+1}\right\|^{2}$ and $\left\|\tilde{e}_{C+1}^{\, T-K+2}\right\|^{2}$ holds:
$$
    \|e_{C+1}\|^{2} \,= \min _{\beta} \,\| e_C -\sum_{i=0}^{T-K+2} {\beta}_{{C+1}}^i p_{C+1}^i \|^{2}  \, \leq  \| {e}_{C} - \sum^{T-K+2}_{i=0} \tilde{\beta}_{C+1}^{\, i} p_{C+1}^{\, i} \|^{2} \, =  \|\tilde{e}_{C+1}^{\, T-K+2}\|^{2}.
$$
    }
    \revise{Besides,} the intermediate prediction error sequence is monotonically decreasing as\reviseS{:}
\begin{align*}
        & \|\tilde{e}_{C+1}^{\, i+1}\|^2-\|\tilde{e}_{C+1}^{\, i}\|^2 \\
={}    & \sum_{h=1}^{H}
\left(
\langle \tilde{e}_{C+1}^{\, i,h}-\tilde{\beta}_{C+1}^{\, i+1,h} p_{C+1}^{i+1}
,
\tilde{e}_{C+1}^{\, i,h}-\tilde{\beta}_{C+1}^{\, i+1,h} p_{C+1}^{i+1} \rangle
-
\langle \tilde{e}_{C+1}^{\, i,h}, \tilde{e}_{C+1}^{\, i,h} \rangle
\right)                                                              \\
={}    & \sum_{h=1}^{H}
\left(
\langle \tilde{\beta}_{C+1}^{\, i+1} p_{C+1}^{i+1}
,
\tilde{\beta}_{C+1}^{\, i+1} p_{C+1}^{i+1} \rangle
-
2 \langle \tilde{e}_{C+1}^{\, i,h} , \tilde{\beta}_{C+1}^{\, i+1} p_{C+1}^{i+1} \rangle
\right)                                                              \\
={}    & \sum_{h=1}^{H}
\left(
- {\langle \tilde{e}_{C+1}^{\, i,h}, p_{C+1}^{\, i+1} \rangle}^2 / \left\|p_{C+1}^{\, i+1}\right\|^{2}
\right)                                                              \\
\leq{} & 0 .
\end{align*}
    \revise{And the inequality between $\left\|\tilde{e}_{C+1}^{\, 0}\right\|^{2}$ and $\left\|e_{C}\right\|^{2}$ can be proven by:
\begin{align*}
& \|\tilde{e}_{C+1}^{\,0}\|^2-\|{e}_{C}\|^2 \\
={}    & \sum_{h=1}^{H}
\left(
\langle {e}_{C}^{\, h}-\tilde{\beta}_{C+1}^{\, 0,h} p_{C+1}^{0}
,
{e}_{C}^{\, h}-\tilde{\beta}_{C+1}^{\, 0,h} p_{C+1}^{0} \rangle
-
\langle {e}_{C}^{\, h}, {e}_{C}^{\, h} \rangle
\right)                                                              \\
={}    & \sum_{h=1}^{H}
\left(
\langle \tilde{\beta}_{C+1}^{\, 0,h} p_{C+1}^{0}
,
\tilde{\beta}_{C+1}^{\, 0,h} p_{C+1}^{0} \rangle
-
2 \langle {e}_{C}^{\, h} , \tilde{\beta}_{C+1}^{\, 0,h} p_{C+1}^{0} \rangle
\right)                                                              
\\
={}    & \sum_{h=1}^{H}
\left(
- {\langle {e}_{C}^{\, h}, p_{C+1}^{\, 0} \rangle}^2 / \left\|p_{C+1}^{\, 0}\right\|^{2}
\right)                                                              \\
\leq{} & 0 .
\end{align*}}
    Then, the convergence of ESM-CNN can be \revise{guaranteed} by\revise{:}
    $$
        \|e_{C+1}\|^2  \; \leq \|\tilde{e}_{C+1}^{\, T-K+2}\|^2 \; \leq \|\tilde{e}_{C+1}^{\,0}\|^2 \; \leq \|{e}_{C}\|^2 .
    $$
\end{proof}
There are three major benefits from conducting the error-feedback calculation individually, rather than globally updating all parameters of the fully connected layer.
The overhead calculation of updating all weights and biases of fully connected layer is more expensive, especially in the situation that a number of filters have been generated.
\revise{Besides, as the random CNN successively adds the filters, a progressively increasing number of features are involved in the construction process, making the least squares method encounter \reviseS{the} ill-posed problem and lose accuracy~\cite{vogel2002computational}.}
Furthermore, the error-feedback fully connected layers use the information \revise{of} prediction error in the last construction process, which is indeed to steadily compensate the prediction error at each process and reduces the uncertainty caused by randomness.

\subsection{Filter Selection Strategy}~\label{subsec:fs}
As per the convolution filter selection problem, we propose a greedy based method to select the random filters with multiple sizes in the construction process of ESM-CNN, such that the different temporal patterns can be extracted.

Assume an ESM-CNN with single convolutional layer has been constructed with $C$ filters, as expressed in \autoref{eq:sccnn}.
Then, following the uniform distribution $[-\lambda, \lambda]$, a batch of subsampled filter candidates are randomly generated $\{p_{{C+1},s} \}^S_{s=1}$ with the corresponding convolutional and pooling filter sizes  $\{(K_{m,s}, K_{p,s})\}^S_{s=1}$.
To evaluate the filter candidates, the filter score $\Delta_{{C+1, s}}$ which represents the prediction promotion after adding the filter candidate $p_{C+1, s}$ is proposed and calculated by\reviseS{:}
\begin{align}
    \Delta_{{C+1},s} & = \| e_{C+1,s} \|^2 - \| e_{C} \|^2 \notag                                                                        \\
    {}               & = \|\ e_C -\sum_{i=0}^{T-K^{\prime}+2} \beta_{C+1, s}^i p_{C+1, s}^i \|^2 -  \| e_{C} \|^2 \label{eq:filterScore}
\end{align}
where $K^{\prime} = K_{m,s} + K_{p,s}$, $p_{C+1, s}$ is computed as \autoref{eq:pool}, and $\beta_{C+1, s}$ is solved as \autoref{eq:scupdate}.

Based on the filter score, the best subsampled filter $p_{C+1}^{*}$ that achieves the maximum promotion is selected from the filter candidates, which is\reviseS{:}
\begin{equation}\label{eq:scselection}
    p_{C+1}^{*} = \argmax\limits_{p_{{C+1},s}} \{ \Delta_{{C+1},s}, s = 1,\ldots,S \}.
\end{equation}
After that, the parameters of the corresponding error-feedback fully connected layer are calculated by \autoref{eq:scupdate}.
Through this filter selection strategy, a variety of temporal patterns are step-wisely extracted for ESM-CNN.

\begin{figure*}[!ht]
    \includegraphics[width = \textwidth]{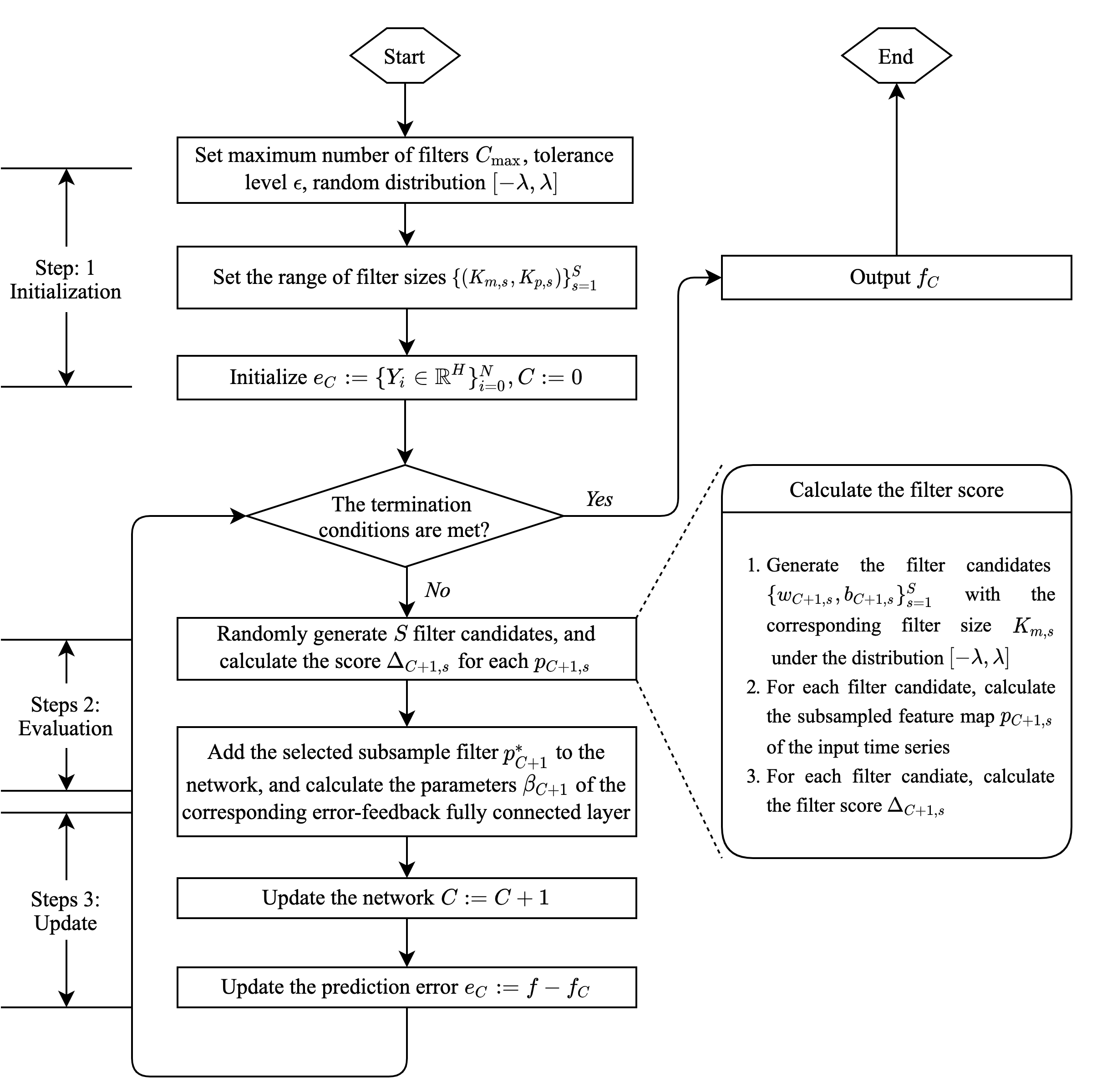}
    \caption{\label{fig:flowchart} Flowchart of the proposed ESM-CNN modeling strategy.}
\end{figure*}
\subsection{Implementation of ESM-CNN}
This subsection details the implementation of the proposed ESM-CNN strategy, as illustrated in \autoref{alg:esc} and~\autoref{fig:flowchart}.
Considering the filter selections, the major operations of ESM-CNN modeling strategy consist of \reviseS{the} initialization, evaluation, updat\reviseS{ing}, and a preprocessing of normalization.
\revise{
For data preprocessing, the data sets are first scaled to standard normal distribution.
\begin{enumerate}
    \item[1)] {
\textbf{Initialization.}
Consider that a time series data $D=\left\{\left(X_{i}, Y_{i}\right) \in\left(\mathbb{R}^{T} \times \mathbb{R}^{H}\right)\right\}_{i=1}^{N}$ is composed of $N$ samples with $T$ observations $[x_1,\dots, x_T]^{\trans}$ for forecasting the next $H$-steps-ahead observations $[x_{T+1},\ldots,x_{T+H}]^{\trans}$.
The proposed ESM-CNN is initialized with several basic configurations, including
the initial prediction error $e_0 $ that is defined as the forecasting target $e_0 = [x_{T+1},\ldots,x_{T+H}]^{\trans}$,
the maximum number of filters $C_{\max}$,
the expected tolerance $\epsilon$,
the random distribution $[-\lambda,\lambda]$,
and the candidate convolution filter sizes as well as the candidate pooling filter sizes $\{(K_{m,s}, K_{p,s})\}^S_{s=1}$.
Concerning \reviseS{the selection of }those configurations, it is yet a challenging model selection problem.
In this work, the expected tolerance $\epsilon$ is set as $0$ to further explore the forecasting performance of the proposed ESM-CNN, the maximum number of filters $C_{\max}$ is set as $100$ to ensure the convergence of the modeling. Besides, the distribution $[-\lambda,\lambda]$ is set as $[-0.5,0.5]$ and the candidate convolution filter sizes are set as $ \{K_{m,s}\} ^S_{s=1} = \{T/3, T/4, T/5, T/6\}$ through our experimental experience.
Following the analysis in~\cite{zhao2017convolutional},
the average pooling with $K_p=3$ is adopted in the experiments.
}
\item[2)] {
\textbf{Evaluation.}
After the initialization, the proposed ESM-CNN steps into the constructing process with no initialized filter in the convolutional layer as $C := 0$.
To select the convolutional filters in the constructing process, a greedy based filter selection strategy is proposed to evaluate and select the random filters with different filter sizes.
In each iteration $C$, we first build a filter candidate set wherein the candidate filters are randomly generated following the distribution $[-\lambda,\lambda]$.
For each candidate filter, we construct a candidate architecture by adding it to the network, then calculate its subsampled feature map\reviseS{s} $\{ p_{C+1,s}\}^S_{s=1}$ and obtain the corresponding score $\{\Delta_{{C+1},s}\}^S_{s=1}$ by \autoref{eq:filterScore}.
After that, we evaluate the candidate filters and select the best one which yields the most significant prediction promotion.
}
\item[3)] {
\textbf{Update.}
Following the evaluation, we calculate the parameters of the error-feedback fully connected layer that is linked with the selected filter by \autoref{eq:scupdate}. 
Then, the selected convolution filter and the corresponding error-feedback fully connected layer are configured into the network, widening the convolutional layer as $C:=C+1$.
Meanwhile, the predict\reviseS{ion} error is steadily decreasing and updated as $e_{C} := f - f_{C}$.
Once the termination conditions are satisfied, a powerful CNN is efficiently crafted for time series forecasting.
}
\end{enumerate}
}

\begin{algorithm}[!ht]
    \renewcommand{\algorithmicrequire}{\textbf{Input:}}
    \renewcommand{\algorithmicensure}{\textbf{Output:}}
    \caption{ESM-CNN Modeling Algorithm}
    \label{alg:esc}
    \begin{algorithmic}[1]
        \REQUIRE {
        time series dataset $D$\reviseS{,} \\
        maximum number of convolution filters $C_{\max}$\reviseS{,} \\
        tolerance level $\epsilon$\reviseS{,} \\
        random distribution $[-\lambda, \lambda]$\reviseS{,} \\
        convolution and pooling filter sizes $\{(K_{m,s}, K_{p,s})\}^S_{s=1}$.
        }
        \ENSURE{
        subsampled convolution filters $\{ p_c \}^C_{c=1}$ ,\\
        corresponding error-feedback fully connected layers $\{ \beta_c\}^C_{c=1}$, where the weight is $[\beta_c^{1}, \dots , \beta_c^{T-K+2}]$ and the bias is $\beta_c^{0}$.
        }
        \STATE Initialize $e_0 := \{ Y_i \in \mathbb{R}^{H}\}^N_{i=1}$, $C := 0$.
        \WHILE{$ C+1 \leq C_{\max}$ and $\| e_C \|_F \geq \epsilon $}

        \STATE Randomly generate $S$ 1D filter candidates following the distribution $[ -\lambda, \lambda]$ and the corresponding convolution as well as the pooling filter size $(K_{m,s}, K_{p,s})$.
        \STATE Calculate the subsampled feature maps $\{ p_{C+1,s}\}^S_{s=1}$ and the corresponding scores $\{\Delta_{{C+1},s}\}^S_{s=1}$.
        \STATE $ p_{C+1}^{*}=\argmax \{ \Delta_{C+1,1},\ldots, \Delta_{C+1,S} \}$ is selected and added to $f_C$.

        \STATE Calculate the parameters of the error-feedback fully connected layer linked with the newly added filter $[\beta_{C+1}^0, \dots, \beta_{C+1}^{T-K+2}]$.
        \STATE Update the number of filters $C:=C+1$.
        \STATE Update the current prediction error $e_{C} := f - f_{C}$.
        \ENDWHILE
        \STATE \textbf{return:} Subsampled convolution filters $\{p_1, \dots, p_C \}$ ,
        and parameters of the corresponding error-feedback fully connected layer $\{ \beta_1,\dots,\beta_C \}$\reviseS{.}
    \end{algorithmic}
\end{algorithm}

\section{Experimental Design \label{sec:design}}

\subsection{Dataset Description}
To evaluate the performance of the proposed ESM-CNN modeling strategy and the counterparts in terms of forecasting accuracy,
one synthetic time series dataset, i.e., \textit{first order autoregression} (AR1), as well as \revise{two} real-world dataset, i.e., \textit{bitcoin price} (BTC) and \textit{influenza-like illness} (ILI), are used for experiments.

The \textit{first order autoregression} time series are recognized as benchmark time series that have been commonly used and reported by a number of studies related to time series modeling and forecasting~\cite{qi2008trend,crone2016feature}.
As an example of stochastic trend series that exhibits complex and chaotic behavior, AR1 is synthesized in \autoref{eq:ar}:
\begin{equation}
\vspace{-0.5em}
\label{eq:ar}
    x_t = \alpha + x_{t-1} + \varepsilon_t,
\end{equation}
where $\alpha = 0.01$, $\varepsilon_t \sim i.i.d.~\mathrm{U}(-0.25,0.25)$, and 500 observed values are simulated.

\revise{The \textit{bitcoin price} constitutes a significant part of the economics and block chain markets. Through bitcoin price forecasting, the investors can be supported for making proper decisions. However, the bitcoin price behaves hardly-predictable fluctuations, making accurate prediction of bitcoin price forecasting be an important and challenging task.
In this real-world forecasting task, we concern two-hourly bitcoin to U.S. dollar close price, and 2181 observed values from May 2020 to March 2021 are drawn from the tradeview service.\footnote{https://www.tradingview.com/symbols/BTCUSD/}}

The \textit{influenza-like illness} not only keeps seriously threatening public health but also makes it more challenging for defending against coronaries disease 2019 (COVID-19) since the ILI shares widely common symptoms with COVID-19. 
Thus, it is crucial to provide accurate influenza prediction to support the public health administrations, making ILI prediction be selected in this study.
In this case, the ILI data is extracted from the Chinese National Influenza Center's website,\footnote{http://www.chinaivdc.cn/cnic/zyzx/lgzb/} and consists of 533 weekly ILI percentage values in southern China from January 2010 to March 2020.

Both \revise{three} datasets are used for evaluating the performance of the proposed ESM-CNN modeling strategy and the counterparts.
Each time series is split into three parts for training, validation and testing following the common relation of 0.64, 0.16 and 0.2.
The embedding dimension $T$ of AR1, BTC, and ILI are 15, \revise{24 (two days), }and 26 (half year) respectively. 



\begin{table}[!ht]
    \centering
    \footnotesize
    \caption{The summary information of the forecasting tasks. \label{tab:data}}
    \begin{tabularx}{\textwidth}{cYYY}
        \toprule
        Dataset             & AR1  & BTC   & ILI     \\
        \midrule
        Observed values     & 500  & 2181   & 533     \\
        Embedding dimension & 15    & 24  & 26      \\
        Prediction horizon\revise{s}  & 1, 3, 6 & 1, 3, 6 & 1, 4, 8 \\
        \bottomrule
    \end{tabularx}

\end{table}
We examine one-step-ahead and multi-step-ahead predictions to justify the performance of the proposed modeling strategies over different horizons.
For multi-step-ahead forecasting, the multiple-input multiple-outputs strategy~\cite{bontempi2008long}, often advocated in standard time series textbooks and reviews, is implemented.
As shown in ~\autoref{tab:data}, the multi-step-ahead horizons for the AR1 dataset are set as 3 and 6.
\revise{The forecasting horizons for BTC dataset are selected as six hours and half day (3 two-hours and 6 two-hours).}
The prediction horizons for ILI dataset are set as about one month and two months (4 weeks and 8 weeks).

\subsection{Counterparts Selection}
\subsubsection{RVFL, IELM and SCN}
To compare the stochastic filters with random hidden neurons, \reviseS{RVFL\cite{igelnik1995stochastic}, IELM\cite{huang2006universal} and SCN\cite{wang2017stochastic}} are selected as counterparts.
The parameters of RVFL, IELM and SCN (i.e., distributions of randomly generating neurons, tolerance level) are kept consistent with ESM-CNN to control the impact variables in the comparisons.
And the pre-defined number of hidden neurons in RVFL is kept same with the maximum number of hidden neurons and filters in IELM, SCN and ESM-CNN.

\subsubsection{GS-CNN, DeepAR and CLSTM}
Besides above mentioned random MLPs, CNN, recurrent neural network (RNN), and convolutional recurrent neural network (CRNN) are also selected as strong competitors.
We implement a single convolutional layer CNN \revise{by using} grid-search approach to select the network architecture, and name it as GS-CNN. 
We implement the RNN with DeepAR~\cite{salinasDeepAR2020} which is built based on an autoregression long short term memory network (LSTM) and shows state-of-the-art performance on many forecasting tasks.
Besides GS-CNN and DeepAR, we implement a convolutional long short term memory network~\cite{livieris2020cnn} which couples the CNN and LSTM as CRNN, and denote this model as CLSTM.
The hyper-parameters (i.e., optimization method, learning rate, and training epochs) of GS-CNN, DeepAR, and CLSTM are set and fine-tuned following trial and error fashion.

\subsubsection{ES-CNN and Stoc-CNN}
Furthermore, we perform an ablation study to verify the necessity of filter selection strategy and error-feedback mechanism in ESM-CNN.
We remove the filter selection strategy and maintain the error-feedback in ESM-CNN, making an ablative counterpart as error-feedback stochastic convolutional neural network (ES-CNN).
Furthermore, we remove the both error-feedback and filter selection strategies of ESM-CNN, making another ablative counterpart as vanilla stochastic convolutional neural network \reviseS{(Stoc-CNN)\cite{yu2019impact}}.

As for controlling variables in ablation experiments, the parameters of Stoc-CNN (i.e., distribution of randomly generating filters, number of filters) follow the same settings of ESM-CNN.
And the maximum number of filters in ESM-CNN and ES-CNN are also kept consistent.

The codes and experiment\revise{al} logs are published in Github repository,\footnote{https://github.com/XinzeZhang/TimeSeriesForecasting-torch} including all details of the \revise{models used} in this study.
Through these counterparts, we can explore and justify the effectiveness of ESM-CNN, which is the highlight of our work.

\subsection{Accuracy Measure}
To evaluate the forecasting accuracy of the proposed ESM-CNN modeling strategy and the selected counterparts from various aspects (i.e., percentage error and numerical error), three widely used statistic measures are selected, including mean absolute percentage error (MAPE), symmetric mean absolute percentage error (SMAPE), and root mean square error (RMSE) \cite{bao2013pso,guo2016robust,zhao2017deep}.
\begin{equation}
	\label{eq:mape}
	MAPE = \frac1N \sideset{}{_{i=1}^N} \sum \abs{\frac{y_{i} - \hat  y_{i}}{y_i}}.
\end{equation}
\begin{equation}
	\label{eq:smape}
	SMAPE = \frac1N \sideset{}{_{i=1}^N} \sum \abs{\frac{y_{i} - \hat  y_{i}}{y_i + \hat  y_{i}}}.
\end{equation}
\begin{equation}
	\label{eq:rmse}
	RMSE = \sqrt{\frac1N \sideset{}{_{i=1}^N} \sum ({y_{i} - \hat  y_{i}})^2 }.
\end{equation}

Here $y_i$ and $\hat y_i$ stand for the real value and the predicted value respectively.

\subsection{Experimental Procedure}
For the selected time series datasets, each of the datasets is \revise{first scaled with z-score method to map onto standard normal distribution,\footnote{https://scikit-learn.org/stable/modules/preprocessing.html\#preprocessing-scaler}} and split into a training, a validation set, and a testing set. 
The proposed ESM-CNN modeling strategy and the counterparts are fitted as well as cross-validated with the training set and the validation set.
\revise{For preprocessed data, we convert the outputs back to their original scales.}
Then, the models \revise{can be estimated and directly compared} for each forecasting horizons with the MAPE, SMAPE, and RMSE measures over all datasets.

Finally, we execute each model 20 times using CUDA 10.1 accelerated Pytorch framework\footnote{https://pytorch.org/} on a Ubuntu 20.04 server with Intel 8700K CPU and Nvidia GTX 1070 GPU,
and the averaged performance is reported in the next section.
\reviseS{Furthermore, upon the requests from reviewers, eight more real-world datasets and three more linear competing models were selected to extend the present experiment, and the results are provided in \ref{app:exp}.}

\section{Results and discussions}
\subsection{Comparison on Prediction Accuracy \label{subsec:acc}}
The prediction performance of the proposed ESM-CNN, the state-of-art training base counterparts (GS-CNN, DeepAR and CLSTM) and random based counterparts (RVFL, IELM and SCN) as well as ablative counterparts (Stoc-CNN and ES-CNN) on all datasets are respectively shown in \revise{\autoref{tab:results_mape}, \autoref{tab:results_smape}, and \autoref{tab:results_rmse}}, examined in terms of three accuracy measures (MAPE, SMPAE, and RMSE).
From the third column to the last column, the mean measure values and the standard deviation (in the bracket) are listed.
For each row of the statistics, the entry with the smallest value is in boldface and marked with an asterisk, and the second smallest value is set in boldface type.

\begin{table*}[!t]
	\centering
    \footnotesize
	\caption{The average MAPE of ESM-CNN and the counterparts. \label{tab:results_mape}}
	\resizebox{\textwidth}{!}{
	\begin{tabular}{ccccccccccc}
		\toprule
		\multirow{2}{*}{Dataset} & \multirow{2}{*}{$H$} & \multicolumn{3}{c}{Training} & \multicolumn{3}{c}{Random} & \multicolumn{2}{c}{Ablation} & Ours                                                                                                                                     \\ 
		\cmidrule(l){3-5} \cmidrule(l){6-8} \cmidrule(lr){9-10} \cmidrule(lr){11-11}
		                      &   & \multicolumn{1}{c}{GS-CNN}          & \multicolumn{1}{c}{DeepAR}          & \multicolumn{1}{c}{CLSTM }          & \multicolumn{1}{c}{RVFL}              & \multicolumn{1}{c}{IELM}            & \multicolumn{1}{c}{SCN}             & \multicolumn{1}{c}{Stoc-CNN}            & \multicolumn{1}{c}{ES-CNN}                  & \multicolumn{1}{c}{ESM-CNN}                         \\
		\midrule
\multirow{6}{*}{AR1} & \multirow{2}{*}{1} & 1.06e-01   & 1.08e-01     & 8.88e-02            & 7.61e-01   & 9.84e-02   & 6.50e-02   & 6.10e+01   & \textbf{4.53e-02}   & \textbf{3.49e-02}\s   \\ \cmidrule(l){3-11} 
                     &                    & (7.38e-03) & (2.26e-02)   & (1.42e-02)          & (4.74e-01) & (1.47e-02) & (2.41e-02) & (2.52e+01) & \textbf{(5.81e-03)} & \textbf{(5.01e-04)} \\ \cmidrule(l){3-11} 
                     & \multirow{2}{*}{3} & 1.23e-01   & 1.64e-01     & 1.20e-01            & 1.17e+00   & 1.07e-01   & 9.53e-02   & 6.34e+01   & \textbf{6.41e-02}   & \textbf{5.28e-02}\s   \\
                     &                    & (7.15e-03) & (3.46e-02)   & (1.34e-02)          & (5.85e-01) & (1.22e-02) & (1.97e-02) & (2.13e+01) & \textbf{(5.21e-03)} & \textbf{(1.77e-03)} \\ \cmidrule(l){3-11} 
                     & \multirow{2}{*}{6} & 1.48e-01   & 2.17e-01     & 1.24e-01            & 1.23e+00   & 1.26e-01   & 1.16e-01   & 1.54e+02   & \textbf{8.92e-02}   & \textbf{7.70e-02}\s   \\
                     &                    & (6.41e-03) & (3.72e-02)   & (1.84e-02)          & (7.78e-01) & (1.74e-02) & (2.63e-02) & (8.02e+01) & \textbf{(5.23e-03)} & \textbf{(2.32e-03)} \\ \cmidrule(l){2-11} 
\multirow{6}{*}{BTC} & \multirow{2}{*}{1} & 6.27e-02   & 2.26e-01     & 4.09e-01            & 1.20e+00   & 1.85e-01   & 1.11e-01   & 2.14e+02   & \textbf{4.62e-02}   & \textbf{2.70e-02}\s   \\
                     &                    & (6.37e-03) & (3.09e-02)   & (2.34e-02)          & (7.24e-01) & (2.98e-02) & (2.45e-02) & (1.90e+02) & \textbf{(1.82e-02)} & \textbf{(4.98e-04)} \\ \cmidrule(l){3-11} 
                     & \multirow{2}{*}{3} & 6.43e-02   & 1.98e-01     & 4.41e-01            & 1.13e+00   & 1.84e-01   & 1.29e-01   & 2.86e+02   & \textbf{4.73e-02}   & \textbf{2.96e-02}\s   \\
                     &                    & (6.63e-03) & (3.80e-02)   & (2.44e-02)          & (4.82e-01) & (2.94e-02) & (2.62e-02) & (2.33e+02) & \textbf{(1.76e-02)} & \textbf{(3.67e-04)} \\ \cmidrule(l){3-11} 
                     & \multirow{2}{*}{6} & 6.49e-02   & 2.35e-01     & 4.61e-01            & 1.38e+00   & 1.84e-01   & 1.40e-01   & 3.32e+02   & \textbf{4.84e-02}   & \textbf{3.26e-02}\s   \\
                     &                    & (6.28e-03) & (4.20e-02)   & (9.60e-03)          & (6.57e-01) & (3.06e-02) & (2.54e-02) & (3.00e+02) & \textbf{(1.68e-02)} & \textbf{(2.65e-04)} \\ \cmidrule(l){2-11} 
\multirow{6}{*}{ILI} & \multirow{2}{*}{1} & 1.28e-01   & 1.14e-01     & \textbf{9.50e-02}   & 3.03e-01   & 1.38e-01   & 1.15e-01   & 2.48e+00   & 1.00e-01            & \textbf{8.93e-02}\s   \\
                     &                    & (2.30e-03) & (8.92e-03)   & \textbf{(2.64e-03)} & (5.17e-02) & (8.14e-03) & (9.50e-03) & (3.72e-01) & (9.59e-03)          & \textbf{(3.10e-03)} \\ \cmidrule(l){3-11} 
                     & \multirow{2}{*}{4} & 1.52e-01   & 1.59e-01     & \textbf{1.26e-01}   & 3.53e-01   & 1.58e-01   & 1.53e-01   & 2.33e+00   & 1.33e-01            & \textbf{1.25e-01}\s   \\
                     &                    & (1.27e-03) & (7.46e-03)   & \textbf{(1.89e-03)} & (3.88e-02) & (5.43e-03) & (9.04e-03) & (2.25e-01) & (4.64e-03)          & \textbf{(3.95e-03)} \\ \cmidrule(l){3-11} 
                     & \multirow{2}{*}{8} & 1.70e-01   & 1.82e-01     & \textbf{1.45e-01}\s   & 3.67e-01   & 1.74e-01   & 1.86e-01   & 2.38e+00   & 1.62e-01            & \textbf{1.53e-01}   \\
                     &                    & (7.76e-04) & (8.70e-03)   & \textbf{(3.17e-03)} & (4.19e-02) & (4.81e-03) & (1.11e-02) & (3.15e-01) & (4.07e-03)          & \textbf{(3.33e-03)} \\ 
		\bottomrule
	\end{tabular}}
\end{table*}
\begin{table*}[!t]
	\centering
    \footnotesize
	\caption{The average SMAPE of ESM-CNN and the counterparts. \label{tab:results_smape}}
	\resizebox{\textwidth}{!}{
	\begin{tabular}{ccccccccccc}
		\toprule
		\multirow{2}{*}{Dataset} & \multirow{2}{*}{$H$} & \multicolumn{3}{c}{Training} & \multicolumn{3}{c}{Random} & \multicolumn{2}{c}{Ablation} & Ours                                                                                                                                     \\ 
		\cmidrule(l){3-5} \cmidrule(l){6-8} \cmidrule(lr){9-10} \cmidrule(lr){11-11}
		                      &   & \multicolumn{1}{c}{GS-CNN}          & \multicolumn{1}{c}{DeepAR}          & \multicolumn{1}{c}{CLSTM }          & \multicolumn{1}{c}{RVFL}              & \multicolumn{1}{c}{IELM}            & \multicolumn{1}{c}{SCN}             & \multicolumn{1}{c}{Stoc-CNN}            & \multicolumn{1}{c}{ES-CNN}                  & \multicolumn{1}{c}{ESM-CNN}                         \\
		\midrule
\multirow{6}{*}{AR1} & \multirow{2}{*}{1} & 5.74e-02   & 5.81e-02     & 4.74e-02            & 3.02e+00   & 5.30e-02   & 3.40e-02   & 1.21e+00   & \textbf{2.33e-02}   & \textbf{1.76e-02}\s   \\
                     &                    & (4.33e-03) & (1.28e-02)   & (8.42e-03)          & (6.74e+00) & (8.80e-03) & (1.35e-02) & (5.13e-01) & \textbf{(3.16e-03)} & \textbf{(3.11e-04)} \\ \cmidrule(l){3-11} 
                     & \multirow{2}{*}{3} & 6.73e-02   & 9.17e-02     & 6.61e-02            & 8.61e+00   & 5.76e-02   & 5.11e-02   & 1.50e+00   & \textbf{3.36e-02}   & \textbf{2.72e-02}\s   \\
                     &                    & (4.23e-03) & (2.10e-02)   & (8.15e-03)          & (1.07e+01) & (7.44e-03) & (1.15e-02) & (1.36e+00) & \textbf{(2.89e-03)} & \textbf{(1.04e-03)} \\ \cmidrule(l){3-11} 
                     & \multirow{2}{*}{6} & 8.28e-02   & 1.25e-01     & 6.82e-02            & 6.36e+00   & 6.97e-02   & 6.28e-02   & 1.52e+00   & \textbf{4.78e-02}   & \textbf{4.05e-02}\s   \\
                     &                    & (3.84e-03) & (2.41e-02)   & (1.17e-02)          & (6.59e+00) & (1.06e-02) & (1.62e-02) & (1.73e+00) & \textbf{(3.03e-03)} & \textbf{(1.53e-03)} \\ \cmidrule(l){2-11} 
\multirow{6}{*}{BTC} & \multirow{2}{*}{1} & 3.26e-02   & 1.29e-01     & 2.60e-01            & 5.04e+00   & 1.04e-01   & 5.93e-02   & 1.12e+00   & \textbf{2.37e-02}   & \textbf{1.35e-02}\s   \\
                     &                    & (3.49e-03) & (2.02e-02)   & (1.87e-02)          & (7.64e+00) & (1.87e-02) & (1.44e-02) & (2.20e-01) & \textbf{(9.73e-03)} & \textbf{(2.90e-04)} \\ \cmidrule(l){3-11} 
                     & \multirow{2}{*}{3} & 3.35e-02   & 1.12e-01     & 2.87e-01            & 2.85e+00   & 1.04e-01   & 7.05e-02   & 1.09e+00   & \textbf{2.43e-02}   & \textbf{1.48e-02}\s   \\
                     &                    & (3.65e-03) & (2.38e-02)   & (1.97e-02)          & (4.01e+00) & (1.85e-02) & (1.54e-02) & (1.26e-01) & \textbf{(9.39e-03)} & \textbf{(2.10e-04)} \\ \cmidrule(l){3-11} 
                     & \multirow{2}{*}{6} & 3.38e-02   & 1.36e-01     & 3.03e-01            & 2.69e+00   & 1.04e-01   & 7.68e-02   & 1.05e+00   & \textbf{2.48e-02}   & \textbf{1.63e-02}\s   \\
                     &                    & (3.47e-03) & (2.76e-02)   & (8.17e-03)          & (4.68e+00) & (1.92e-02) & (1.52e-02) & (6.03e-02) & \textbf{(8.99e-03)} & \textbf{(1.44e-04)} \\ \cmidrule(l){2-11} 
\multirow{6}{*}{ILI} & \multirow{2}{*}{1} & 6.91e-02   & 6.02e-02     & \textbf{4.86e-02}   & 1.28e-01   & 7.49e-02   & 6.06e-02   & 9.65e-01   & 5.20e-02            & \textbf{4.60e-02}\s   \\
                     &                    & (1.48e-03) & (4.37e-03)   & \textbf{(1.52e-03)} & (1.73e-02) & (4.74e-03) & (5.13e-03) & (5.61e-01) & (5.09e-03)          & \textbf{(1.42e-03)} \\ \cmidrule(l){3-11} 
                     & \multirow{2}{*}{4} & 8.40e-02   & 8.43e-02     & \textbf{6.71e-02}   & 1.64e-01   & 8.83e-02   & 8.24e-02   & 1.39e+00   & 7.13e-02            & \textbf{6.65e-02}\s   \\
                     &                    & (7.92e-04) & (3.34e-03)   & \textbf{(1.09e-03)} & (1.67e-02) & (3.64e-03) & (5.04e-03) & (4.77e-01) & (2.42e-03)          & \textbf{(1.97e-03)} \\ \cmidrule(l){3-11} 
                     & \multirow{2}{*}{8} & 9.79e-02   & 1.00e-01     & \textbf{8.11e-02}\s   & 2.55e-01   & 1.01e-01   & 1.04e-01   & 1.60e+00   & 9.07e-02            & \textbf{8.52e-02}   \\
                     &                    & (5.25e-04) & (3.68e-03)   & \textbf{(1.99e-03)} & (1.63e-01) & (3.15e-03) & (6.38e-03) & (5.97e-01) & (2.25e-03)          & \textbf{(1.75e-03)} \\ 
		\bottomrule
	\end{tabular}}
\end{table*}
\begin{table*}[!t]
	\centering
    \footnotesize
	\caption{The average RMSE of ESM-CNN and the counterparts. \label{tab:results_rmse}}
	\resizebox{\textwidth}{!}{
	\begin{tabular}{ccccccccccc}
		\toprule
		\multirow{2}{*}{Dataset} & \multirow{2}{*}{$H$} & \multicolumn{3}{c}{Training} & \multicolumn{3}{c}{Random} & \multicolumn{2}{c}{Ablation} & Ours                                                                                                                                     \\ 
		\cmidrule(l){3-5} \cmidrule(l){6-8} \cmidrule(lr){9-10} \cmidrule(lr){11-11}
		                      &   & \multicolumn{1}{c}{GS-CNN}          & \multicolumn{1}{c}{DeepAR}          & \multicolumn{1}{c}{CLSTM }          & \multicolumn{1}{c}{RVFL}              & \multicolumn{1}{c}{IELM}            & \multicolumn{1}{c}{SCN}             & \multicolumn{1}{c}{Stoc-CNN}            & \multicolumn{1}{c}{ES-CNN}                  & \multicolumn{1}{c}{ESM-CNN}                         \\
		\midrule
\multirow{6}{*}{AR1}  & \multirow{2}{*}{1} & 5.53e-01   & 5.08e-01     & 4.89e-01            & 3.21e+00   & 5.54e-01   & 3.34e-01   & 4.36e+02   & \textbf{2.28e-01}   & \textbf{1.55e-01}\s   \\
                      &                    & (3.60e-02) & (9.60e-02)   & (8.93e-02)          & (1.85e+00) & (7.81e-02) & (1.13e-01) & (2.01e+02) & \textbf{(3.43e-02)} & \textbf{(3.66e-03)} \\ \cmidrule(l){3-11} 
                      & \multirow{2}{*}{3} & 6.30e-01   & 7.66e-01     & 6.67e-01            & 5.15e+00   & 5.80e-01   & 4.98e-01   & 4.43e+02   & \textbf{3.22e-01}   & \textbf{2.57e-01}\s   \\
                      &                    & (3.26e-02) & (1.30e-01)   & (6.71e-02)          & (2.26e+00) & (6.50e-02) & (8.72e-02) & (1.95e+02) & \textbf{(2.43e-02)} & \textbf{(1.04e-02)} \\ \cmidrule(l){3-11} 
                      & \multirow{2}{*}{6} & 7.40e-01   & 9.96e-01     & 6.79e-01            & 5.59e+00   & 6.71e-01   & 6.00e-01   & 1.07e+03   & \textbf{4.61e-01}   & \textbf{3.94e-01}\s   \\
                      &                    & (2.55e-02) & (1.34e-01)   & (9.10e-02)          & (3.27e+00) & (7.89e-02) & (1.27e-01) & (5.37e+02) & \textbf{(2.71e-02)} & \textbf{(1.62e-02)} \\ \cmidrule(l){2-11} 
\multirow{6}{*}{BTC}  & \multirow{2}{*}{1} & 3.87e+03   & 1.15e+04     & 2.08e+04            & 6.74e+04   & 1.05e+04   & 6.52e+03   & 1.33e+07   & \textbf{2.75e+03}   & \textbf{1.59e+03}\s   \\
                      &                    & (3.62e+02) & (1.52e+03)   & (1.15e+03)          & (3.63e+04) & (1.60e+03) & (1.41e+03) & (1.11e+07) & \textbf{(1.07e+03)} & \textbf{(2.39e+01)} \\ \cmidrule(l){3-11} 
                      & \multirow{2}{*}{3} & 3.97e+03   & 1.02e+04     & 2.24e+04            & 7.11e+04   & 1.04e+04   & 7.55e+03   & 1.92e+07   & \textbf{2.82e+03}   & \textbf{1.74e+03}\s   \\
                      &                    & (3.76e+02) & (1.91e+03)   & (1.15e+03)          & (2.70e+04) & (1.58e+03) & (1.40e+03) & (1.43e+07) & \textbf{(1.03e+03)} & \textbf{(2.01e+01)} \\ \cmidrule(l){3-11} 
                      & \multirow{2}{*}{6} & 4.00e+03   & 1.22e+04     & 2.33e+04            & 8.82e+04   & 1.04e+04   & 8.09e+03   & 2.30e+07   & \textbf{2.89e+03}   & \textbf{1.92e+03}\s   \\
                      &                    & (3.54e+02) & (2.12e+03)   & (4.55e+02)          & (3.93e+04) & (1.64e+03) & (1.33e+03) & (1.96e+07) & \textbf{(9.79e+02)} & \textbf{(1.51e+01)} \\  \cmidrule(l){2-11}                     
\multirow{6}{*}{ILI} & \multirow{2}{*}{1} & 8.45e-01   & 7.82e-01     & 7.10e-01            & 1.74e+00   & 8.88e-01   & 7.14e-01   & 1.50e+01   & \textbf{6.81e-01}   & \textbf{6.06e-01}\s   \\
                      &                    & (1.95e-02) & (1.87e-02)   & (2.52e-02)          & (3.17e-01) & (4.41e-02) & (3.11e-02) & (2.50e+00) & \textbf{(5.37e-02)} & \textbf{(8.14e-03)} \\ \cmidrule(l){3-11} 
                      & \multirow{2}{*}{4} & 1.00e+00   & 9.95e-01     & 9.07e-01            & 1.96e+00   & 1.04e+00   & 9.09e-01   & 1.39e+01   & \textbf{8.65e-01}   & \textbf{8.14e-01}\s   \\
                      &                    & (9.40e-03) & (1.52e-02)   & (1.71e-02)          & (2.49e-01) & (3.56e-02) & (2.52e-02) & (1.54e+00) & \textbf{(2.35e-02)} & \textbf{(8.75e-03)} \\ \cmidrule(l){3-11} 
                      & \multirow{2}{*}{8} & 1.16e+00   & 1.16e+00     & \textbf{1.04e+00}   & 2.00e+00   & 1.18e+00   & 1.11e+00   & 1.43e+01   & 1.07e+00            & \textbf{1.03e+00}\s   \\
                      &                    & (5.03e-03) & (2.11e-02)   & \textbf{(1.76e-02)} & (2.32e-01) & (2.07e-02) & (2.54e-02) & (2.03e+00) & (1.07e-02)          & \textbf{(6.63e-03)} \\ 

		\bottomrule
	\end{tabular}}
\end{table*}

The results of comparison between ESM-CNN and the state-of-art competitors in Table \ref{tab:results_mape}, \autoref{tab:results_smape}, and \autoref{tab:results_rmse} lead to the following conclusions.
\begin{enumerate}
    \item[1)] {As the random neural network, ESM-CNN keeps promising accuracy and small variance on both artificial and real-world datasets, showing that ESM-CNN is robust and stable to practical applications.
          }
    \item[2)] {
          In comparison with trained methods,
          GS-CNN outperforms DeepAR and CLSTM on the \revise{BTC} dataset while CLSTM outperforms GS-CNN and DeepAR on the \revise{AR1 and ILI} datasets,
          which shows that the convolutional architecture can be effectiveness for time series forecasting.
    }
    \item[3\revise{)}] {
          Compared with trained methods, taking the advantage of the convolutional architecture, ESM-CNN achieves the best RMSE on \revise{all} datasets as well as all three prediction horizons and achieves almost the same percentage performance with CLSTM on ILI dataset, which demonstrates the effectiveness of ESM-CNN on characterizing the time series.
          }
    \item[4\revise{)}]  {
        In comparison with random based counterparts, the incrementally constructed random MLPs (IELM and SCN) outperform the deterministic constructed RVFL, leading to the same conclusion in the \revise{related} studies~\cite{huang2006universal,wang2017stochastic} and showing the effectiveness of the incrementally constructive method.
    }
    \item[5\revise{)}] {
        Compared with random based counterparts, ESM-CNN also outperforms RVFL, IELM, and SCN on \revise{all} datasets, showing the strong predictive power of ESM-CNN for time series forecasting.
          }
\end{enumerate}

The ablation study in \revise{\autoref{tab:results_mape}, \autoref{tab:results_smape}, and \autoref{tab:results_rmse}} shows the results as follows.
\begin{enumerate}
    \item[1\revise{)}] {
        The Stoc-CNN, which uses purely deterministic random filters and the least square method to solve the output weights, achieves the worst performance on all tasks, indicating that the \revise{ill-posed problem} occurs after globally configuring the output layer of the random CNN with a number of random filters.
    \item[2\revise{)}] {
        The ES-CNN, which incrementally constructs the random convolutional layer and computes the output weights based on the error-feedback strategy, \revise{achieves a huge improvement of forecasting performance on Stoc-CNN,} and shows the promising performance on \revise{all} dataset, demonstrating the effectiveness of the incrementally constructive method for random CNN.
    }
    \item[3\revise{)}] {
        Based the comparison between ES-CNN and ESM-CNN, despite a narrow margin of the performance occurs on AR1 dataset, ESM-CNN surpasses ES-CNN in all experiments on \revise{real-world datasets (BTC and ILI)}, demonstrating the necessity of introducing the filter selection strategy to construct the \revise{random} convolutional layer \revise{in} ESM-CNN.
    }
    } 
\end{enumerate}

\begin{figure*}[!t]
    \centering
    \begin{minipage}[b]{0.32\textwidth}

    \end{minipage}

    \begin{minipage}[b]{0.32\textwidth}
        \includegraphics[width = 0.95\textwidth]{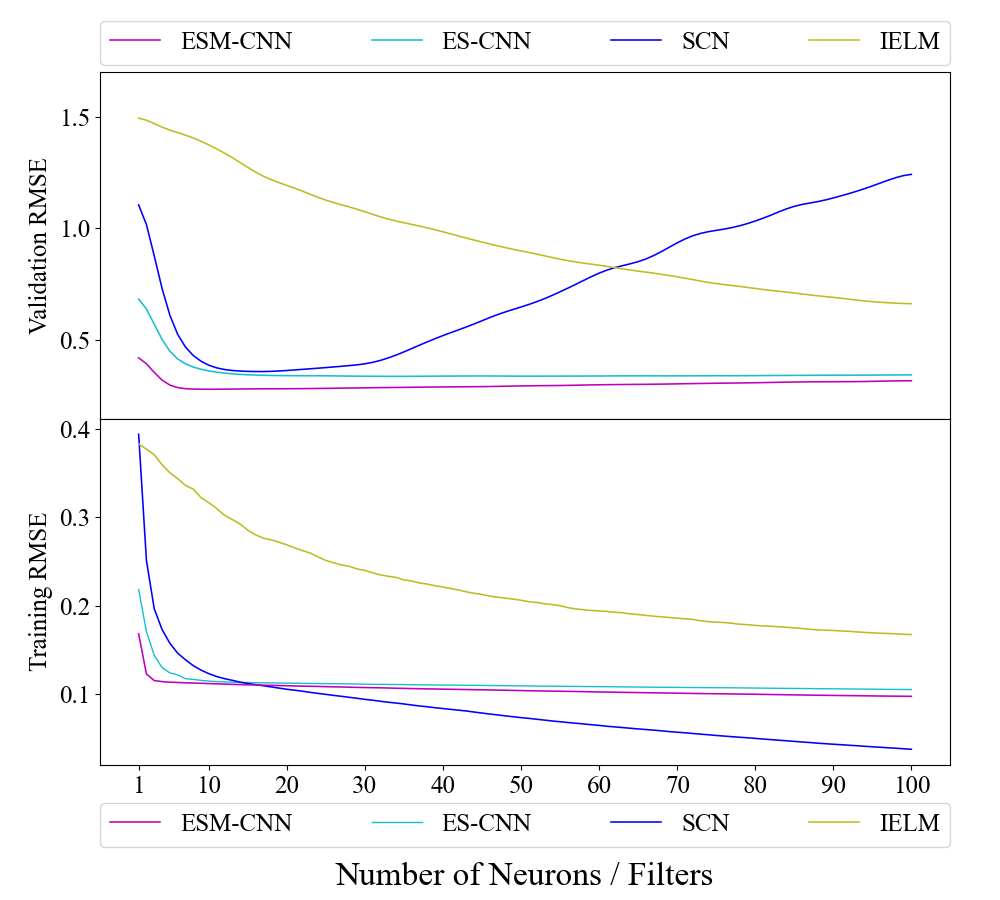}
        \subcaption{\label{fig:silih1} ILI, $H = 1$ }
    \end{minipage}
    \begin{minipage}[b]{0.32\textwidth}
        \includegraphics[width = 0.95\textwidth]{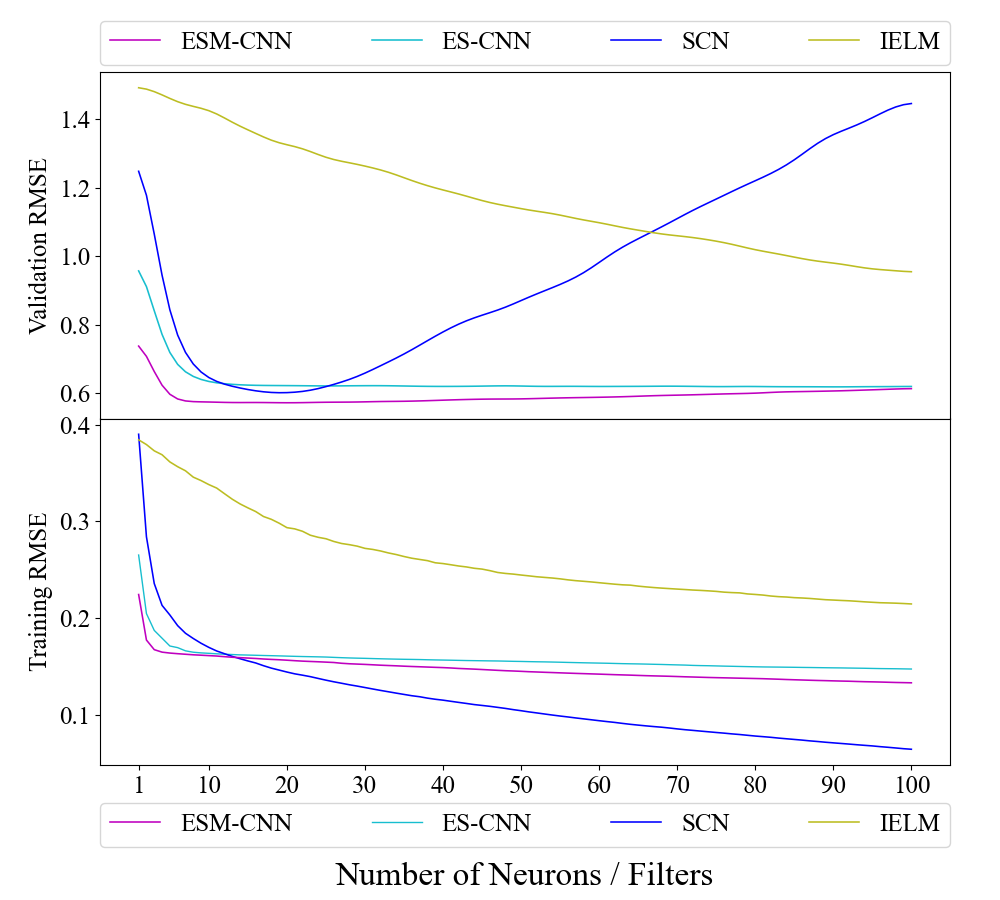}
        \subcaption{\label{fig:silih4} ILI, $H = 4$ }
    \end{minipage}
    \begin{minipage}[b]{0.32\textwidth}
        \includegraphics[width = 0.95\textwidth]{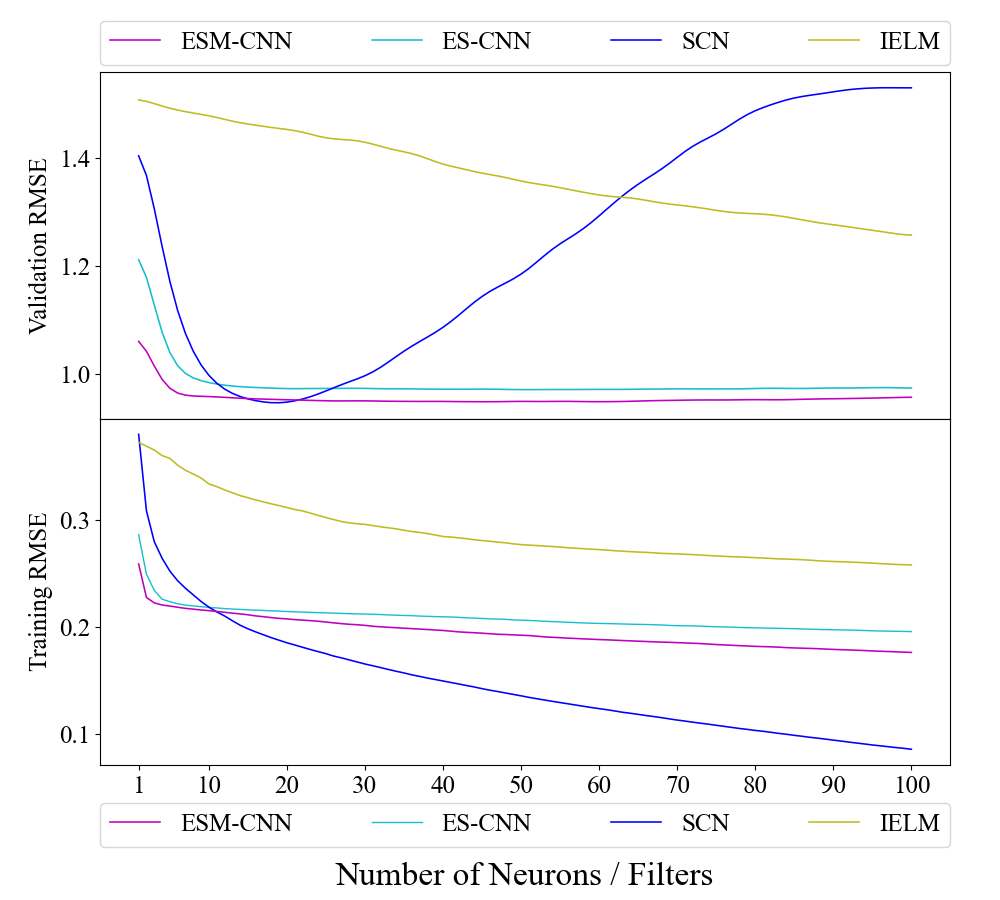}
        \subcaption{\label{fig:silih8} ILI, $H = 8$ }
    \end{minipage}

    \caption{\label{fig:converge} The average RMSE of incrementally constructed random neural networks on ILI dataset.}
\end{figure*}

\subsection{Comparison on Convergence}
We further examine the convergence \revise{among} the incrementally constructive modeling strategies (i.e., IELM, SCN, ES-CNN, and ESM-CNN).
The averaged training and validation RMSE are plotted as the functions of the generated neurons or filters, which are shown in \autoref{fig:converge}.
The results lead to the following conclusions.
\begin{enumerate}
    \item[1)] {
        Although we have examined \revise{various} prediction horizons for each dataset, general conclusions are consistent with the \revise{different} prediction horizons for all datasets.
          }
    \item[2)] {
        Despite that SCN outperforms other models in terms of the rapid decreasing of RMSE on all training sets, it shows obvious overfitting on all validation sets.
        In contrast to SCN, IELM does not overfit on the training datasets but obviously converges slower than other models.
          }
    \item[3)] {
          Via the error-feedback fully connected layer, ESM-CNN and ES-CNN outperform SCN and IELM on all validation sets, demonstrating the robustness and stability when considering the error-feedback strategy to construct random CNN for time series forecasting.
          }
    \item[4)] {
          Comparing ESM-CNN with ES-CNN, despite that the convergence curves of the two strategies are close to each other, there are still remarkable findings. 
          ESM-CNN steadily converges to a slightly but existentially lower RMSE on the training processes and exhibits better performance on the validation sets, which suggests the value of filter selection strategy.
          }
    \item[5)]{
          For the convergence of the ESM-CNN and ES-CNN modeling strategies, the most significant improvements come from the early 10 constructive iterations.
          A possible explanation is that, the benefits from the error-feedback fully connected layer gradually vanishes as the number of filters increases, which gives a guidance that the smaller computation overheads may be enough to bring efficiency improvements \revise{on} implementing ESM-CNN to time series prediction.
          }
\end{enumerate}

Here we try to provide an explanation to the overfitting issue of SCN and Stoc-CNN as well as the robustness of ESC-CNN and ES-CNN.
The SCN and Stoc-CNN both globally compute all parameters of the fully connected output layer with the least square method, which makes the output parameters excessively complex and easily overfit on the noisy data of the training set.
By contrast, the error-feedback modeling strategy of ESM-CNN and ES-CNN individually calculates the parameters of the newly added neuron at the error-feedback fully connected layer to compensate prediction error in \revise{the} previous construction iteration. 
Thus, the influence of the noisy data is naturally reduced during the construction process, and the output weights are configured with \revise{a} smaller norm than SCN and Stoc-CNN, enabling ESC-CNN and ES-CNN with higher robustness and stability.

\begin{table*}[!t]
    \centering
    \footnotesize
    \caption{The average running time (seconds) of each model on both two datasets. \label{tab:time}}
    \resizebox{\textwidth}{!}{\begin{tabular}{ccrrrrrrrrr}
        \toprule
        \multirow{2}{*}{Methods}   & \multirow{2}{*}{Models} & \multicolumn{3}{c}{AR1} & \multicolumn{3}{c}{BTC} & \multicolumn{3}{c}{ILI}                                           \\\cmidrule(lr){3-5} \cmidrule(lr){6-8}\cmidrule(lr){9-11}
                          &                         & \multicolumn{1}{c}{1} & \multicolumn{1}{c}{3} & \multicolumn{1}{c}{6} & \multicolumn{1}{c}{1} & \multicolumn{1}{c}{3} & \multicolumn{1}{c}{6} & \multicolumn{1}{c}{1} & \multicolumn{1}{c}{4} & \multicolumn{1}{c}{8} \\ \midrule        
        
\multirow{3}{*}{Training} & GS-CNN                  & 74.65                 & 76.95                 & 78.55                 & 93.70                 & 96.45                 & 99.40                 & 67.05                 & 70.25                 & 73.00                 \\
                          & DeepAR                  & 225.25                & 238.95                & 257.70                & 223.75                & 251.10                & 268.65                & 256.80                & 279.10                & 275.30                \\
                          & CLSTM                   & 150.45                & 154.80                & 160.25                & 143.00                & 141.50                & 142.25                & 167.40                & 170.80                & 175.30                \\
\specialrule{0em}{1.5pt}{1.5pt}                          
\multirow{3}{*}{Random}   & RVFL                    & \multicolumn{1}{c}{-} & \multicolumn{1}{c}{-} & \multicolumn{1}{c}{-} & \multicolumn{1}{c}{-}  & \multicolumn{1}{c}{-}  & \multicolumn{1}{c}{-}  & \multicolumn{1}{c}{-} & \multicolumn{1}{c}{-} & \multicolumn{1}{c}{-} \\
                          & IELM                    & 2.65                  & 2.85                  & 2.80                  & 3.40                  & 3.30                  & 3.20                  & 2.55                  & 2.55                  & 2.40                  \\
                          & SCN                     & 22.85                 & 57.40                 & 107.75                & 21.10                 & 58.00                 & 111.40                & 16.10                 & 49.60                 & 90.95                 \\
\specialrule{0em}{1.5pt}{1.5pt}                          
\multirow{2}{*}{Ablation} & Stoc-CNN                & \multicolumn{1}{c}{-} & \multicolumn{1}{c}{-} & \multicolumn{1}{c}{-} & \multicolumn{1}{c}{-}  & \multicolumn{1}{c}{-}  & \multicolumn{1}{c}{-}  & \multicolumn{1}{c}{-} & \multicolumn{1}{c}{-} & \multicolumn{1}{c}{-} \\
                          & ES-CNN                  & 6.50                  & 6.40                  & 6.50                  & 6.35                  & 6.50                  & 6.45                  & 6.45                  & 6.50                  & 6.50                  \\
\specialrule{0em}{1.5pt}{1.5pt}                          
Our                       & ESM-CNN                 & 9.65                  & 9.65                  & 9.70                  & 7.10                  & 7.25                  & 7.25                  & 7.15                  & 7.15                  & 7.20                  \\ \bottomrule
    \end{tabular}}
\end{table*}
Furthermore, the average running time (seconds) of each model on \revise{all} datasets are shown in \autoref{tab:time}, 
The results show that, comparing with the trained models (GS-CNN, DeepAR, and CLSTM) as well as SCN, ESM-CNN and ES-CNN shows the competitive performance with a very short time.
Besides, the proposed greedy based filter selection strategy in ESM-CNN only consumes a few additional time while bringing the significant improvements as shown in the \autoref{subsec:acc}, demonstrating the efficiency of the ESM-CNN.

\section{Conclusion}
Time series forecasting is a challenging task due to the unsteady and dynamic innate property. 
In this work, we develop an error-feedback stochastic modeling strategy of convolutional neural network (ESM-CNN) for time series forecasting.
Through the error-feedback stochastic modeling and the filter selection, ESM-CNN \revise{exhibits promising predictive power} for time series forecasting with theoretical guarantee.
\revise{The} superiority of the proposed strategies with iteratively adding convolutional filters and configuring the corresponding error-feedback fully connected layer as well as the greedy based filter selection method are demonstrated experimentally.

\revise{
In future work, we would like to try the ESM strategy to recurrent neural network architectures.
Since the proposed ESM strategy widens the hidden layer with an incremental mechanism, we would also adapt this method to a hybrid architecture which consists of convolutional filters and recurrent cells in the same hidden layer, making the model benefit from the convolutional neural architectures and recurrent neural architectures simultaneously.
Besides, we would propose a deep version of ESM strategy to utlize multiple layers to promote the forecasting performance.}

\section{Acknowledgment}
This work was supported by Natural Science Foundation of China under Project Nos. 71571080 and 71871101, and Fundamental Research Fund for Central Universities under Project No.2019kfyXKJC021.

\bibliography{mybibfile,revise}

\newpage
\appendix
\section{Proof of the convergence of ESM-CNN}\label{app:proof}
   The intermediate prediction error sequence $\tilde{e}_{C+1}^{\, 0}, \ldots, \tilde{e}_{C+1}^{\, T-K+2} $ and intermediate parameters $\tilde{\beta}_{C+1}^{\, 0}, \ldots, \tilde{\beta}_{C+1}^{\, T-K+2}$ of the new added error-feedback fully connected layer are introduced as
    \begin{alignat*}{2}
         & \tilde{e}_{C+1}^{\, i+1}       & = & \mspace{18mu} \tilde{e}_{C+1}^{\, i}-\tilde{\beta}_{C+1}^{\, i+1} p_{C+1}^{i+1}, \quad i = 0,\ldots,T-K+1,                                        \\
        \shortintertext{where}
         & \tilde{\beta}_{C+1}^{\, i+1}   & = & \mspace{18mu} [\tilde{\beta}_{C+1}^{\, i+1,1}, \ldots,\tilde{\beta}_{C+1}^{\, i+1,h},\ldots, \tilde{\beta}_{C+1}^{\, i+1,H}],                     \\
         & \tilde{\beta}_{C+1}^{\, i+1,h} & = & \mspace{18mu} \left\langle \tilde{e}_{C+1}^{\, i,h}, p_{C+1}^{\, i+1}\right\rangle /\left\|p_{C+1}^{\, i+1}\right\|^{2} , \quad h= 1, \ldots, H , \\
        \shortintertext{and}
         & \tilde{e}_{C+1}^{\, 0}         & = & \mspace{18mu} e_{C}- \tilde{\beta}_{C+1}^{\, 0} p_{C+1}^{\, 0},                                                                                   \\
         & \tilde{\beta}_{C+1}^{\, 0}   & = & \mspace{18mu} [\tilde{\beta}_{C+1}^{\, 0,1}, \ldots,\tilde{\beta}_{C+1}^{\, 0,h},\ldots, \tilde{\beta}_{C+1}^{\, 0,H}],                     \\
         & \tilde{\beta}_{C+1}^{\, 0,h} & = & \mspace{18mu} \left\langle {e}_{C}^{\, h}, p_{C+1}^{\, 0}\right\rangle /\left\|p_{C+1}^{\, 0}\right\|^{2} , \quad h= 1, \ldots, H , \\
         \shortintertext{making}
         & \tilde{e}_{C+1}^{\, T-K+2} \,  & = & \mspace{18mu} {e}_{C} - \sum^{T-K+2}_{i=0} \tilde{\beta}_{C+1}^{\, i} p_{C+1}^{\, i}.
    \end{alignat*}

    Since the parameters of the new added error-feedback fully connected layer are calculated by the least square method:
    $$
        \left[\beta_{{C+1}}^0, \ldots, \beta_{{C+1}}^{T-K+2} \right]=\argmin _{\beta} \,\|e_C -\sum_{i=0}^{T-K+2} \beta_{C+1}^i p_{C+1}^i \|,
    $$
    the basic inequality between $\left\|e_{C+1}\right\|^{2}$ and $\left\|\tilde{e}_{C+1}^{\, T-K+2}\right\|^{2}$ holds:
$$
    \|e_{C+1}\|^{2} \,= \min _{\beta} \,\| e_C -\sum_{i=0}^{T-K+2} {\beta}_{{C+1}}^i p_{C+1}^i \|^{2}  \, \leq  \| {e}_{C} - \sum^{T-K+2}_{i=0} \tilde{\beta}_{C+1}^{\, i} p_{C+1}^{\, i} \|^{2} \, =  \|\tilde{e}_{C+1}^{\, T-K+2}\|^{2}.
$$

    Besides, the intermediate prediction error sequence is monotonically decreasing as:

\begin{align*}
& \|\tilde{e}_{C+1}^{\, i+1}\|^2-\|\tilde{e}_{C+1}^{\, i}\|^2 \\
={}    & \sum_{h=1}^{H}
\left(
\langle \tilde{e}_{C+1}^{\, i,h}-\tilde{\beta}_{C+1}^{\, i+1,h} p_{C+1}^{i+1}
,
\tilde{e}_{C+1}^{\, i,h}-\tilde{\beta}_{C+1}^{\, i+1,h} p_{C+1}^{i+1} \rangle
-
\langle \tilde{e}_{C+1}^{\, i,h}, \tilde{e}_{C+1}^{\, i,h} \rangle
\right)                                                              \\
={}    & \sum_{h=1}^{H}
\left(
\langle \tilde{\beta}_{C+1}^{\, i+1} p_{C+1}^{i+1}
,
\tilde{\beta}_{C+1}^{\, i+1} p_{C+1}^{i+1} \rangle
-
2 \langle \tilde{e}_{C+1}^{\, i,h} , \tilde{\beta}_{C+1}^{\, i+1} p_{C+1}^{i+1} \rangle
\right)                                                              
\\
={}    & \sum_{h=1}^{H}
\left(
\langle 
\frac{\left\langle \tilde{e}_{C+1}^{\, i,h}, p_{C+1}^{\, i+1}\right\rangle}{\left\|p_{C+1}^{\, i+1}\right\|^{2}}  
p_{C+1}^{i+1}
,
\frac{\left\langle \tilde{e}_{C+1}^{\, i,h}, p_{C+1}^{\, i+1}\right\rangle}{\left\|p_{C+1}^{\, i+1}\right\|^{2}}  
p_{C+1}^{i+1}
\rangle
-
2 
\langle 
\tilde{e}_{C+1}^{\, i,h} 
,
\frac{\left\langle \tilde{e}_{C+1}^{\, i,h}, p_{C+1}^{\, i+1}\right\rangle}{\left\|p_{C+1}^{\, i+1}\right\|^{2}}  
p_{C+1}^{i+1} 
\rangle
\right)                                                              
\\
={}    & \sum_{h=1}^{H}
\left(
\frac{{\left\langle \tilde{e}_{C+1}^{\, i,h}, p_{C+1}^{\, i+1}\right\rangle}^2}{\left\|p_{C+1}^{\, i+1}\right\|^{4}}  
\langle 
p_{C+1}^{i+1}
,
p_{C+1}^{i+1}
\rangle
-
2 \frac{\left\langle \tilde{e}_{C+1}^{\, i,h}, p_{C+1}^{\, i+1}\right\rangle}{\left\|p_{C+1}^{\, i+1}\right\|^{2}}  
\langle \tilde{e}_{C+1}^{\, i,h} 
,
p_{C+1}^{i+1} 
\rangle
\right)                                                              
\\
={}    & \sum_{h=1}^{H}
\left(
\frac{{\left\langle \tilde{e}_{C+1}^{\, i,h}, p_{C+1}^{\, i+1}\right\rangle}^2}{\left\|p_{C+1}^{\, i+1}\right\|^{4}}  
{\left\|p_{C+1}^{\, i+1}\right\|^{2}}  
-
2 \frac{{\left\langle \tilde{e}_{C+1}^{\, i,h}, p_{C+1}^{\, i+1}\right\rangle}^2}{\left\|p_{C+1}^{\, i+1}\right\|^{2}}  
\right)                                                              
\\        
={}    & \sum_{h=1}^{H}
\left(
- {\langle \tilde{e}_{C+1}^{\, i,h}, p_{C+1}^{\, i+1} \rangle}^2 / \left\|p_{C+1}^{\, i+1}\right\|^{2}
\right)                                                              \\
\leq{} & 0 .
\end{align*}

    And the inequality between $\left\|\tilde{e}_{C+1}^{\, 0}\right\|^{2}$ and $\left\|e_{C}\right\|^{2}$ can be proven by:

    \begin{align*}
& \|\tilde{e}_{C+1}^{\,0}\|^2-\|{e}_{C}\|^2 \\
={}    & \sum_{h=1}^{H}
\left(
\langle {e}_{C}^{\, h}-\tilde{\beta}_{C+1}^{\, 0,h} p_{C+1}^{0}
,
{e}_{C}^{\, h}-\tilde{\beta}_{C+1}^{\, 0,h} p_{C+1}^{0} \rangle
-
\langle {e}_{C}^{\, h}, {e}_{C}^{\, h} \rangle
\right)                                                              \\
={}    & \sum_{h=1}^{H}
\left(
\langle \tilde{\beta}_{C+1}^{\, 0,h} p_{C+1}^{0}
,
\tilde{\beta}_{C+1}^{\, 0,h} p_{C+1}^{0} \rangle
-
2 \langle {e}_{C}^{\, h} , \tilde{\beta}_{C+1}^{\, 0,h} p_{C+1}^{0} \rangle
\right)                                                              
\\
={}    & \sum_{h=1}^{H}
\left(
\langle 
\frac{\left\langle {e}_{C}^{\, h}, p_{C+1}^{\, 0}\right\rangle}{\left\|p_{C+1}^{\, 0}\right\|^{2}}  
p_{C+1}^{0}
,
\frac{\left\langle {e}_{C}^{\, h}, p_{C+1}^{\, 0}\right\rangle}{\left\|p_{C+1}^{\, 0}\right\|^{2}}  
p_{C+1}^{0}
\rangle
-
2 
\langle 
{e}_{C}^{\, h} 
,
\frac{\left\langle {e}_{C}^{\, h}, p_{C+1}^{\, 0}\right\rangle}{\left\|p_{C+1}^{\, 0}\right\|^{2}}  
p_{C+1}^{0} 
\rangle
\right)                                                              
\\
={}    & \sum_{h=1}^{H}
\left(
\frac{{\left\langle {e}_{C}^{\, h}, p_{C+1}^{\, 0}\right\rangle}^2}{\left\|p_{C+1}^{\, 0}\right\|^{4}}  
\langle 
p_{C+1}^{0}
,
p_{C+1}^{0}
\rangle
-
2 \frac{\left\langle {e}_{C}^{\, h}, p_{C+1}^{\, 0}\right\rangle}{\left\|p_{C+1}^{\, 0}\right\|^{2}}  
\langle {e}_{C}^{\, h} 
,
p_{C+1}^{0} 
\rangle
\right)                                                              
\\
={}    & \sum_{h=1}^{H}
\left(
\frac{{\left\langle {e}_{C}^{\, h}, p_{C+1}^{\, 0}\right\rangle}^2}{\left\|p_{C+1}^{\, 0}\right\|^{4}}  
{\left\|p_{C+1}^{\, 0}\right\|^{2}}  
-
2 \frac{{\left\langle {e}_{C}^{\, h}, p_{C+1}^{\, 0}\right\rangle}^2}{\left\|p_{C+1}^{\, 0}\right\|^{2}}  
\right)                                                              
\\        
={}    & \sum_{h=1}^{H}
\left(
- {\langle {e}_{C}^{\, h}, p_{C+1}^{\, 0} \rangle}^2 / \left\|p_{C+1}^{\, 0}\right\|^{2}
\right)                                                              \\
\leq{} & 0 .
\end{align*}

    Then, the convergence of ESM-CNN can be proven by
    $$
        \|e_{C+1}\|^2  \; \leq \|\tilde{e}_{C+1}^{\, T-K+2}\|^2 \; \leq \|\tilde{e}_{C+1}^{\,0}\|^2 \; \leq \|{e}_{C}\|^2 .
    $$

\setcounter{section}{1}
\section{Additional experiments}\label{app:exp}

In this appendix, we consider eight more real-world datasets to evaluate our method to further strengthen the experiments.
The additional datasets used in this study are as follows:
\begin{itemize}
    \item \textit{Weekly Europe Brent crude oil price} (BRENT-weekly),
    \item \textit{Daily Europe Brent crude oil price} (BRENT-daily),
    \item \textit{Weekly Cushing, OK WTI crude oil price} (WTI-weekly),
    \item \textit{Daily Cushing, OK WTI crude oil price} (WTI-daily),
    \item \textit{Daily close value of S\&P 500 index} (S\&P 500),
    \item \textit{Daily close value of NASDAQ Composite} (NASDAQ),
    \item \textit{Daily close value of Dow Jones Average} (DJI),
    \item \textit{Daily close value of NYSE Composite} (NYSE).
\end{itemize}

\begin{table*}[!ht]
    \centering
    \footnotesize
    \caption{The statistical information of the datasets. \label{tab:app_data}}
    \begin{tabularx}{\textwidth}{lYYYcY}
    \toprule
    Dataset      & Stationarity & Trend & Seasonality &  Period  & Sample size \\ \midrule
    AR1          & \xmark      & 0.97      & 0.09        & -                           & 500         \\
    BTC          & \xmark      & 0.99      & 0.66        & 05/25/2020 $\sim$ 11/23/2020 & 2181        \\
    ILI          & \cmark      & 0.51      & 0.61        & 01/15/2010 $\sim$ 04/15/2020 & 535         \\
    BRENT-weekly & \xmark      & 0.97      & 0.07        & 05/15/1987 $\sim$ 04/30/2021 & 1773        \\
    BRENT-daily  & \xmark      & 0.97      & 0.06        & 05/20/1987 $\sim$ 05/03/2021 & 8620        \\
    WTI-weekly   & \xmark      & 0.96      & 0.08        & 01/03/1986 $\sim$ 04/30/2021 & 1844        \\
    WTI-daily    & \xmark      & 0.96      & 0.07        & 01/02/1986 $\sim$ 05/03/2021 & 8904        \\
    S\&P 500           & \xmark      & 0.99      & 0.40        & 12/31/2009 $\sim$ 11/15/2017 & 1984        \\
    NASDAQ       & \xmark      & 0.99      & 0.27        & 12/31/2009 $\sim$ 11/15/2017 & 1984        \\
    DJI          & \xmark      & 0.99      & 0.40        & 12/31/2009 $\sim$ 11/15/2017 & 1984        \\
    NYSE         & \xmark      & 0.98      & 0.45        & 12/31/2009 $\sim$ 11/15/2017 & 1984        \\ \bottomrule
    \end{tabularx}
    \end{table*}
The Brent crude oil price datasets\footnote{https://www.eia.gov/dnav/pet/hist/RBRTEd.htm} and WTI curde oil price datasets\footnote{ https://www.eia.gov/dnav/pet/hist/rwtcW.htm} are respectively drawn from websites of the U.S. Energy Information Administration, And the other four financial datasets, i.e., S\&P 500, NASDAQ, DJI, and NYSE, are collected from UCI Machine Learning Repository\footnote{https://archive.ics.uci.edu/ml/machine-learning-databases/00554/}.
\autoref{tab:app_data} shows the statistical information of the additional datasets, where the stationarity was estimated with Augmented Dickey Fuller (ADF) test\footnote{https://www.statsmodels.org/stable/generated/statsmodels.tsa.stattools.adfuller.html}, the trend and seasonality are measured with its strength\footnote{https://otexts.com/fpp2/seasonal-strength.html}.
For the column of stationarity, ``\cmark'' denotes the dataset is stationary, otherwise non-stationary.
For the column of trend and seasonality, a bigger score represents a stronger strength of trend or seasonality.

As for the counterparts selection, we further consider three widely used statistical methods, i.e., naive forecasting (Naive), Autoregressive Integrated Moving Average (ARIMA), and Holt’s Winters Seasonal Exponential Smoothing (Holt), to validate our method. The embedding dimension $T$ of weekly additional datasets, i.e., BRENT-weekly and WTI-weekly, are both set as 26 (half year), which are kept the same with weekly ILI dataset. The embedding dimension $T$ of daily additional datasets, i.e., BRENT-daily, WTI-daily, S\&P 500, NASDAQ, DJI, and NYSE, are set as 30 (six weeks).

The results of comparative experiments on all datasets are provided in \autoref{tab:app_mape}, \autoref{tab:app_smape}, and \autoref{tab:app_rmse} respectively, which lead to the same conclusion as in the main text.

\newcommand{\rot}[1]{\rotatebox[origin=c]{90}{{\parbox[c]{1cm}{\centering #1}}}}

\begin{table*}[!ht]
    \centering
    \footnotesize
    \caption{The average MAPE of ESM-CNN and the counterparts. \label{tab:app_mape}}
    \resizebox{\textwidth}{!}{
        \begin{tabular}{cccccccccccccc}
            \toprule
            \multirow{2}{*}{Dataset} & \multirow{2}{*}{H} & \multicolumn{3}{c}{Statistical} & \multicolumn{3}{c}{Training} & \multicolumn{3}{c}{Random} & \multicolumn{2}{c}{Ablation} & Ours                                                                                                                                \\ \cmidrule(l){3-5} \cmidrule(lr){6-8} \cmidrule(lr){9-11} \cmidrule(lr){12-13} \cmidrule(lr){14-14}
                                     &                    & Naive                           & ARIMA                        & Holt                       & CNN                          & DeepAR            & CLSTM             & RVFL     & IELM     & SCN               & Stoc-CNN & ES-CNN            & ESM-CNN           \\ \cmidrule(l){1-14}
            \multirow{3}{*}{{AR1}}    & 1                  & 2.54e-01                        & 5.95e-01                     & 5.94e-01                   & 1.06e-01                     & 1.08e-01          & 8.88e-02          & 7.61e-01 & 9.84e-02 & 6.50e-02          & 6.10e+01 & \textbf{4.53e-02} & \textbf{3.49e-02}\s \\\cmidrule(l){3-14}
                                     & 3                  & 2.62e-01                        & 6.00e-01                     & 5.98e-01                   & 1.23e-01                     & 1.64e-01          & 1.20e-01          & 1.17e+00 & 1.07e-01 & 9.53e-02          & 6.34e+01 & \textbf{6.41e-02} & \textbf{5.28e-02}\s \\\cmidrule(l){3-14}
                                     & 6                  & 2.68e-01                        & 6.08e-01                     & 6.05e-01                   & 1.48e-01                     & 2.17e-01          & 1.24e-01          & 1.23e+00 & 1.26e-01 & 1.16e-01          & 1.54e+02 & \textbf{8.92e-02} & \textbf{7.70e-02}\s \\\cmidrule(l){2-14}
            \multirow{3}{*}{BTC}     & 1                  & 2.55e-01                        & 3.17e-01                     & 3.17e-01                   & 6.27e-02                     & 2.26e-01          & 4.09e-01          & 1.20e+00 & 1.85e-01 & 1.11e-01          & 2.14e+02 & \textbf{4.62e-02} & \textbf{2.70e-02}\s \\\cmidrule(l){3-14}
                                     & 3                  & 2.54e-01                        & 3.18e-01                     & 3.18e-01                   & 6.43e-02                     & 1.98e-01          & 4.41e-01          & 1.13e+00 & 1.84e-01 & 1.29e-01          & 2.86e+02 & \textbf{4.73e-02} & \textbf{2.96e-02}\s \\\cmidrule(l){3-14}
                                     & 6                  & 2.53e-01                        & 3.21e-01                     & 3.20e-01                   & 6.49e-02                     & 2.35e-01          & 4.61e-01          & 1.38e+00 & 1.84e-01 & 1.40e-01          & 3.32e+02 & \textbf{4.84e-02} & \textbf{3.26e-02}\s \\\cmidrule(l){2-14}
            \multirow{3}{*}{ILI}     & 1                  & 1.94e-01                        & 5.70e-01                     & 5.94e-01                   & 1.28e-01                     & 1.14e-01          & \textbf{9.50e-02} & 3.03e-01 & 1.38e-01 & 1.15e-01          & 2.48e+00 & 1.00e-01          & \textbf{8.93e-02}\s \\\cmidrule(l){3-14}
                                     & 4                  & 2.51e-01                        & 5.39e-01                     & 5.90e-01                   & 1.52e-01                     & 1.59e-01          & \textbf{1.26e-01} & 3.53e-01 & 1.58e-01 & 1.53e-01          & 2.33e+00 & 1.33e-01          & \textbf{1.25e-01}\s \\\cmidrule(l){3-14}
                                     & 8                  & 3.07e-01                        & 5.34e-01                     & 5.98e-01                   & 1.70e-01                     & 1.82e-01          & \textbf{1.45e-01}\s & 3.67e-01 & 1.74e-01 & 1.86e-01          & 2.38e+00 & 1.62e-01          & \textbf{1.53e-01} \\\cmidrule(l){2-14}
            \multirowcell{3}{BRENT-\\weekly} & 1                  & 5.02e-01                        & 1.09e-01                     & 1.17e-01                   & 1.60e-01                     & \textbf{4.37e-02} & 8.25e-02          & 5.88e-02 & 1.57e-01 & 5.74e-02          & 1.93e+00 & 4.57e-02          & \textbf{3.97e-02}\s \\\cmidrule(l){3-14}
                                     & 4                  & 4.97e-01                        & 2.13e-01                     & 2.37e-01                   & 1.78e-01                     & 9.32e-02          & 1.15e-01          & 1.19e-01 & 1.73e-01 & 8.67e-02          & 2.87e+00 & \textbf{7.78e-02} & \textbf{7.49e-02}\s \\\cmidrule(l){3-14}
                                     & 8                  & 5.10e-01                        & 3.26e-01                     & 3.69e-01                   & 1.82e-01                     & 1.77e-01          & 1.85e-01          & 1.81e-01 & 2.12e-01 & 1.43e-01          & 3.70e+00 & \textbf{1.14e-01} & \textbf{1.11e-01}\s \\\cmidrule(l){2-14}
            \multirowcell{3}{BRENT-\\daily} & 1                  & 4.80e-01                        & 8.47e-02                     & 8.92e-02                   & 7.84e-02                     & \textbf{1.93e-02}\s & 3.51e-02          & 2.14e-02 & 1.36e-01 & 2.01e-02          & 5.06e-02 & 2.02e-02          & \textbf{1.99e-02} \\\cmidrule(l){3-14}
                                     & 5                  & 4.82e-01                        & 2.26e-01                     & 2.45e-01                   & 9.37e-02                     & 3.48e-02          & 4.44e-02          & 3.92e-02 & 1.42e-01 & 3.50e-02          & 8.78e-02 & \textbf{3.42e-02} & \textbf{3.37e-02}\s \\\cmidrule(l){3-14}
                                     & 10                 & 4.88e-01                        & 3.99e-01                     & 4.34e-01                   & 9.50e-02                     & 5.18e-02          & 5.50e-02          & 5.51e-02 & 1.49e-01 & \textbf{5.10e-02} & 1.29e-01 & 5.37e-02          & \textbf{4.63e-02}\s \\\cmidrule(l){2-14}
            \multirowcell{3}{WTI-\\weekly}    & 1                  & 4.71e-01                        & 9.99e-02                     & 1.10e-01                   & 2.26e-01                     & \textbf{5.49e-02}\s & 1.01e-01          & 7.10e-02 & 2.34e-01 & 6.33e-02          & 2.10e+00 & 6.27e-02          & \textbf{5.77e-02} \\\cmidrule(l){3-14}
                                     & 4                  & 4.78e-01                        & 2.06e-01                     & 2.40e-01                   & 2.47e-01                     & 1.11e-01          & 1.45e-01          & 1.34e-01 & 2.53e-01 & 1.06e-01          & 3.11e+00 & \textbf{9.46e-02} & \textbf{8.64e-02}\s \\\cmidrule(l){3-14}
                                     & 8                  & 5.11e-01                        & 3.18e-01                     & 3.71e-01                   & 2.46e-01                     & 1.72e-01          & 2.16e-01          & 2.06e-01 & 2.81e-01 & 1.35e-01          & 3.70e+00 & \textbf{1.32e-01} & \textbf{1.24e-01}\s \\\cmidrule(l){2-14}
            \multirowcell{3}{WTI-\\daily}    & 1                  & 4.44e-01                        & 7.82e-02                     & 8.33e-02                   & 9.30e-02                     & 2.11e-02          & 3.62e-02          & 2.15e-02 & 1.45e-01 & 2.14e-02          & 7.24e-02 & \textbf{2.08e-02} & \textbf{2.06e-02}\s \\\cmidrule(l){3-14}
                                     & 5                  & 4.49e-01                        & 2.05e-01                     & 2.21e-01                   & 9.90e-02                     & \textbf{3.50e-02}\s & 4.65e-02          & 4.14e-02 & 1.51e-01 & 3.67e-02          & 1.22e-01 & 3.62e-02          & \textbf{3.57e-02} \\\cmidrule(l){3-14}
                                     & 10                 & 4.60e-01                        & 3.57e-01                     & 3.86e-01                   & 9.96e-02                     & 5.21e-02          & 5.76e-02          & 5.93e-02 & 1.58e-01 & \textbf{5.20e-02} & 1.67e-01 & 5.99e-02          & \textbf{4.99e-02}\s \\\cmidrule(l){2-14}
            \multirow{3}{*}{S\&P 500}      & 1                  & 1.50e-01                        & \textbf{5.17e-03}            & 8.26e-03                   & 1.92e-02                     & 1.28e-02          & 4.88e-02          & 8.82e-03 & 3.95e-02 & 1.18e-02          & 1.73e+00 & 5.33e-03          & \textbf{4.06e-03}\s \\\cmidrule(l){3-14}
                                     & 5                  & 1.46e-01                        & 7.15e-03                     & 1.10e-02                   & 2.03e-02                     & 1.87e-02          & 7.20e-02          & 1.91e-02 & 5.20e-02 & 2.20e-02          & 1.09e+01 & \textbf{7.03e-03} & \textbf{6.17e-03}\s \\\cmidrule(l){3-14}
                                     & 10                 & 1.48e-01                        & 8.91e-03                     & 1.32e-02                   & 1.94e-02                     & 7.97e-02          & 5.95e-02          & 2.40e-02 & 4.65e-02 & 3.46e-02          & 1.43e+01 & \textbf{8.71e-03} & \textbf{8.03e-03}\s \\\cmidrule(l){2-14}
            \multirow{3}{*}{NASDAQ}  & 1                  & 1.77e-01                        & 7.16e-03                     & 1.04e-02                   & 2.84e-02                     & 7.19e-03          & 6.86e-02          & 3.65e-02 & 5.78e-02 & 1.62e-02          & 3.85e+01 & \textbf{6.68e-03} & \textbf{5.50e-03}\s \\\cmidrule(l){3-14}
                                     & 5                  & 1.73e-01                        & 1.03e-02                     & 1.40e-02                   & 3.09e-02                     & 2.15e-02          & 9.06e-02          & 3.50e-02 & 6.12e-02 & 2.36e-02          & 6.13e+01 & \textbf{9.12e-03} & \textbf{8.84e-03}\s \\\cmidrule(l){3-14}
                                     & 10                 & 1.75e-01                        & 1.29e-02                     & 1.73e-02                   & 3.05e-02                     & 1.53e-01          & 8.47e-02          & 3.52e-02 & 6.00e-02 & 4.35e-02          & 6.43e+01 & \textbf{1.15e-02} & \textbf{1.13e-02}\s \\\cmidrule(l){2-14}
            \multirow{3}{*}{DJI}     & 1                  & 1.35e-01                        & 5.24e-03                     & 8.14e-03                   & 2.36e-02                     & 1.07e-02          & 4.28e-02          & 1.55e-02 & 4.73e-02 & 7.57e-03          & 6.20e+00 & \textbf{5.06e-03} & \textbf{4.09e-03}\s \\\cmidrule(l){3-14}
                                     & 5                  & 1.31e-01                        & 7.53e-03                     & 1.10e-02                   & 2.54e-02                     & 6.72e-02          & 6.54e-02          & 1.84e-02 & 4.72e-02 & 1.93e-02          & 4.73e+00 & \textbf{7.10e-03} & \textbf{6.71e-03}\s \\\cmidrule(l){3-14}
                                     & 10                 & 1.33e-01                        & 9.55e-03                     & 1.34e-02                   & 2.56e-02                     & 3.74e-02          & 6.64e-02          & 3.60e-02 & 4.70e-02 & 3.89e-02          & 7.16e+00 & \textbf{9.14e-03} & \textbf{9.06e-03}\s \\\cmidrule(l){2-14}
            \multirow{3}{*}{NYSE}    & 1                  & 1.12e-01                        & 5.61e-03                     & 8.91e-03                   & 1.31e-02                     & 9.30e-03          & 1.53e-02          & 6.55e-03 & 2.63e-02 & 9.61e-03          & 6.68e-01 & \textbf{4.58e-03} & \textbf{4.52e-03}\s \\\cmidrule(l){3-14}
                                     & 5                  & 1.09e-01                        & 7.68e-03                     & 1.19e-02                   & 1.44e-02                     & 1.93e-02          & 2.30e-02          & 1.82e-02 & 2.65e-02 & 1.31e-02          & 9.32e-01 & \textbf{6.77e-03} & \textbf{6.69e-03}\s \\\cmidrule(l){3-14}
                                     & 10                 & 1.10e-01                        & 9.24e-03                     & 1.42e-02                   & 1.53e-02                     & 9.84e-02          & 2.89e-02          & 2.60e-02 & 2.69e-02 & 1.81e-02          & 1.45e+00 & \textbf{8.44e-03} & \textbf{8.38e-03}\s \\ \bottomrule
        \end{tabular}}
\end{table*}
\begin{table*}[!ht]
    \centering
    \footnotesize
    \caption{The average SMAPE of ESM-CNN and the counterparts. \label{tab:app_smape}}
    \resizebox{\textwidth}{!}{
        \begin{tabular}{cccccccccccccc}
            \toprule
            \multirow{2}{*}{Dataset} & \multirow{2}{*}{H} & \multicolumn{3}{c}{Statistical} & \multicolumn{3}{c}{Training} & \multicolumn{3}{c}{Random} & \multicolumn{2}{c}{Ablation} & Ours                                                                                                                                \\ \cmidrule(l){3-5} \cmidrule(lr){6-8} \cmidrule(lr){9-11} \cmidrule(lr){12-13} \cmidrule(lr){14-14}
                                     &                    & Naive                           & ARIMA                        & Holt                       & CNN                          & DeepAR            & CLSTM             & RVFL     & IELM     & SCN               & Stoc-CNN & ES-CNN            & ESM-CNN           \\ \cmidrule(l){1-14}
            \multirow{3}{*}{AR1}     & 1                  & 1.33e-01                        & 4.31e-01                     & 4.30e-01                   & 5.74e-02                     & 5.81e-02          & 4.74e-02          & 3.02e+00 & 5.30e-02 & 3.40e-02          & 1.21e+00 & \textbf{2.33e-02} & \textbf{1.76e-02}\s \\ \cmidrule(l){3-14}
                                     & 3                  & 1.37e-01                        & 4.36e-01                     & 4.34e-01                   & 6.73e-02                     & 9.17e-02          & 6.61e-02          & 8.61e+00 & 5.76e-02 & 5.11e-02          & 1.50e+00 & \textbf{3.36e-02} & \textbf{2.72e-02}\s \\ \cmidrule(l){3-14}
                                     & 6                  & 1.46e-01                        & 4.43e-01                     & 4.40e-01                   & 8.28e-02                     & 1.25e-01          & 6.82e-02          & 6.36e+00 & 6.97e-02 & 6.28e-02          & 1.52e+00 & \textbf{4.78e-02} & \textbf{4.05e-02}\s \\ \cmidrule(l){2-14}
            \multirow{3}{*}{BTC}     & 1                  & 1.35e-01                        & 1.90e-01                     & 1.90e-01                   & 3.26e-02                     & 1.29e-01          & 2.60e-01          & 5.04e+00 & 1.04e-01 & 5.93e-02          & 1.12e+00 & \textbf{2.37e-02} & \textbf{1.35e-02}\s \\ \cmidrule(l){3-14}
                                     & 3                  & 1.35e-01                        & 1.91e-01                     & 1.90e-01                   & 3.35e-02                     & 1.12e-01          & 2.87e-01          & 2.85e+00 & 1.04e-01 & 7.05e-02          & 1.09e+00 & \textbf{2.43e-02} & \textbf{1.48e-02}\s \\ \cmidrule(l){3-14}
                                     & 6                  & 1.36e-01                        & 1.93e-01                     & 1.92e-01                   & 3.38e-02                     & 1.36e-01          & 3.03e-01          & 2.69e+00 & 1.04e-01 & 7.68e-02          & 1.05e+00 & \textbf{2.48e-02} & \textbf{1.63e-02}\s \\ \cmidrule(l){2-14}
            \multirow{3}{*}{ILI}     & 1                  & 9.82e-02                        & 4.16e-01                     & 4.47e-01                   & 6.91e-02                     & 6.02e-02          & \textbf{4.86e-02} & 1.28e-01 & 7.49e-02 & 6.06e-02          & 9.65e-01 & 5.20e-02          & \textbf{4.60e-02}\s \\ \cmidrule(l){3-14}
                                     & 4                  & 1.22e-01                        & 3.86e-01                     & 4.44e-01                   & 8.40e-02                     & 8.43e-02          & \textbf{6.71e-02} & 1.64e-01 & 8.83e-02 & 8.24e-02          & 1.39e+00 & 7.13e-02          & \textbf{6.65e-02}\s \\ \cmidrule(l){3-14}
                                     & 8                  & 1.55e-01                        & 3.80e-01                     & 4.54e-01                   & 9.79e-02                     & 1.00e-01          & \textbf{8.11e-02}\s & 2.55e-01 & 1.01e-01 & 1.04e-01          & 1.60e+00 & 9.07e-02          & \textbf{8.52e-02} \\ \cmidrule(l){2-14}
            \multirowcell{3}{BRENT-\\weekly} & 1                  & 3.25e-01                        & 1.00e-01                     & 9.90e-02                   & 6.70e-02                     & \textbf{2.07e-02} & 3.89e-02          & 2.82e-02 & 6.64e-02 & 2.84e-02          & 9.65e-01 & 2.26e-02          & \textbf{1.97e-02}\s \\ \cmidrule(l){3-14}
                                     & 4                  & 9.68e-01                        & 6.22e-01                     & 3.17e-01                   & 7.37e-02                     & 4.17e-02          & 5.37e-02          & 5.42e-02 & 7.15e-02 & 4.19e-02 & 1.14e+00 & \textbf{3.79e-02}          & \textbf{3.67e-02}\s\\ \cmidrule(l){3-14}
                                     & 8                  & 3.68e-01                        & 7.03e-01                     & 4.40e-01                   & 7.74e-02                     & 7.57e-02          & 7.91e-02          & 8.30e-02 & 8.42e-02 & 6.06e-02          & 1.40e+00 & \textbf{5.35e-02} & \textbf{5.25e-02}\s \\ \cmidrule(l){2-14}
            \multirowcell{3}{BRENT-\\daily} & 1                  & 6.55e-01                        & 1.63e-01                     & 1.69e-01                   & 3.67e-02                     & \textbf{9.55e-03}\s & 1.73e-02          & 1.05e-02 & 6.05e-02 & 9.92e-03          & 3.61e-02 & 1.00e-02          & \textbf{9.91e-03} \\ \cmidrule(l){3-14}
                                     & 5                  & 4.22e-01                        & 5.21e-01                     & 4.84e-01                   & 4.31e-02                     & 1.72e-02          & 2.18e-02          & 1.87e-02 & 6.26e-02 & 1.71e-02          & 6.66e-02 & \textbf{1.70e-02} & \textbf{1.67e-02}\s \\ \cmidrule(l){3-14}
                                     & 10                 & 4.17e-01                        & 7.59e-01                     & 6.83e-01                   & 4.38e-02                     & 2.50e-02          & 2.68e-02          & 2.58e-02 & 6.51e-02 & \textbf{2.44e-02} & 1.39e-01 & 2.60e-02          & \textbf{2.28e-02}\s \\ \cmidrule(l){2-14}
            \multirowcell{3}{WTI-\\weekly}   & 1                  & 3.69e-01                        & 7.82e-02                     & 9.42e-02                   & 7.89e-02                     & \textbf{2.06e-02}\s  & 4.17e-02          & 3.08e-02 & 8.89e-02 & 2.48e-02          & 6.86e-01 & 2.52e-02          & \textbf{2.44e-02}\\ \cmidrule(l){3-14}
                                     & 4                  & 9.06e+01                        & 2.10e-01                     & 2.81e-01                   & 8.50e-02                     & 4.00e-02          & 5.86e-02          & 6.84e-02 & 9.40e-02 & 4.13e-02          & 8.43e-01 & \textbf{3.97e-02} & \textbf{3.90e-02}\s \\ \cmidrule(l){3-14}
                                     & 8                  & 3.34e-01                        & 3.80e-01                     & 5.26e-01                   & 8.43e-02                     & 5.92e-02          & 7.93e-02          & 2.88e-01 & 1.00e-01 & \textbf{5.51e-02} & 2.75e+00 & 5.53e-02          & \textbf{5.17e-02}\s \\ \cmidrule(l){2-14}
            \multirowcell{3}{WTI-\\daily}    & 1                  & 8.29e-01                        & 1.34e-01                     & 1.75e-01                   & 4.26e-02                     & 1.05e-02          & 1.78e-02          & 1.06e-02 & 6.45e-02 & 1.06e-02          & 1.17e-01 & \textbf{1.04e-02} & \textbf{1.02e-02}\s \\ \cmidrule(l){3-14}
                                     & 5                  & 3.63e-01                        & 6.75e-01                     & 4.02e-01                   & 4.50e-02                     & \textbf{1.73e-02}\s & 2.27e-02          & 1.95e-02 & 6.67e-02 & 1.79e-02          & 1.10e-01 & 1.76e-02          & \textbf{1.75e-02} \\ \cmidrule(l){3-14}
                                     & 10                 & 4.20e-01                        & 7.36e-01                     & 8.09e-01                   & 4.53e-02                     & 2.51e-02          & 2.78e-02          & 2.73e-02 & 6.94e-02 & \textbf{2.47e-02} & 1.46e-01 & 2.83e-02          & \textbf{2.41e-02}\s \\ \cmidrule(l){2-14}
            \multirow{3}{*}{S\&P 500}      & 1                  & 7.60e-02                        & \textbf{2.59e-03}            & 4.13e-03                   & 9.68e-03                     & 6.47e-03          & 2.53e-02          & 4.41e-03 & 2.03e-02 & 5.97e-03          & 5.23e+00 & 2.66e-03          & \textbf{2.03e-03}\s \\ \cmidrule(l){3-14}
                                     & 5                  & 7.46e-02                        & 3.58e-03                     & 5.50e-03                   & 1.02e-02                     & 9.48e-03          & 3.79e-02          & 9.68e-03 & 2.69e-02 & 1.12e-02          & 4.37e+00 & \textbf{3.50e-03} & \textbf{3.08e-03}\s \\ \cmidrule(l){3-14}
                                     & 10                 & 7.57e-02                        & 4.47e-03                     & 6.61e-03                   & 9.81e-03                     & 4.24e-02          & 3.11e-02          & 1.22e-02 & 2.40e-02 & 1.77e-02          & 7.42e+00 & \textbf{4.34e-03} & \textbf{4.02e-03}\s \\ \cmidrule(l){2-14}
            \multirow{3}{*}{NASDAQ}  & 1                  & 9.04e-02                        & 3.58e-03                     & 5.19e-03                   & 1.44e-02                     & 3.61e-03          & 3.61e-02          & 1.76e-02 & 3.00e-02 & 8.22e-03          & 1.34e+00 & \textbf{3.33e-03} & \textbf{2.75e-03}\s \\ \cmidrule(l){3-14}
                                     & 5                  & 8.91e-02                        & 5.15e-03                     & 7.01e-03                   & 1.57e-02                     & 1.09e-02          & 4.85e-02          & 1.69e-02 & 3.19e-02 & 1.20e-02          & 1.88e+00 & \textbf{4.54e-03} & \textbf{4.42e-03}\s \\ \cmidrule(l){3-14}
                                     & 10                 & 9.08e-02                        & 6.50e-03                     & 8.69e-03                   & 1.55e-02                     & 8.73e-02          & 4.53e-02          & 1.71e-02 & 3.12e-02 & 2.24e-02          & 1.77e+00 & \textbf{5.73e-03} & \textbf{5.66e-03}\s \\ \cmidrule(l){2-14}
            \multirow{3}{*}{DJI}     & 1                  & 6.83e-02                        & 2.62e-03                     & 4.08e-03                   & 1.20e-02                     & 5.36e-03          & 2.21e-02          & 7.66e-03 & 2.45e-02 & 3.80e-03          & 4.52e-01 & \textbf{2.52e-03} & \textbf{2.05e-03}\s \\ \cmidrule(l){3-14}
                                     & 5                  & 6.71e-02                        & 3.78e-03                     & 5.51e-03                   & 1.29e-02                     & 3.58e-02          & 3.44e-02          & 9.04e-03 & 2.44e-02 & 9.78e-03          & 7.47e-01 & \textbf{3.54e-03} & \textbf{3.36e-03}\s \\ \cmidrule(l){3-14}
                                     & 10                 & 6.81e-02                        & 4.80e-03                     & 6.74e-03                   & 1.30e-02                     & 1.92e-02          & 3.50e-02          & 1.73e-02 & 2.43e-02 & 2.01e-02          & 1.60e+00 & \textbf{4.57e-03} & \textbf{4.53e-03}\s \\ \cmidrule(l){2-14}
            \multirow{3}{*}{NYSE}    & 1                  & 5.63e-02                        & 2.81e-03                     & 4.46e-03                   & 6.60e-03                     & 4.67e-03          & 7.73e-03          & 3.28e-03 & 1.33e-02 & 4.83e-03          & 1.12e+00 & \textbf{2.29e-03} & \textbf{2.26e-03}\s \\ \cmidrule(l){3-14}
                                     & 5                  & 5.52e-02                        & 3.84e-03                     & 5.98e-03                   & 7.27e-03                     & 9.77e-03          & 1.17e-02          & 9.21e-03 & 1.35e-02 & 6.59e-03          & 4.99e+00 & \textbf{3.38e-03} & \textbf{3.35e-03}\s \\ \cmidrule(l){3-14}
                                     & 10                 & 5.61e-02                        & 4.64e-03                     & 7.12e-03                   & 7.73e-03                     & 5.49e-02          & 1.48e-02          & 1.33e-02 & 1.37e-02 & 9.13e-03          & 3.56e+00 & \textbf{4.22e-03} & \textbf{4.19e-03}\s \\ \bottomrule
        \end{tabular}}
\end{table*}
\begin{table*}[!ht]
    \centering
    \footnotesize
    \caption{The average RMSE of ESM-CNN and the counterparts. \label{tab:app_rmse}}
    \resizebox{\textwidth}{!}{
        \begin{tabular}{cccccccccccccc}
            \toprule
            \multirow{2}{*}{Dataset} & \multirow{2}{*}{H} & \multicolumn{3}{c}{Statistical} & \multicolumn{3}{c}{Training} & \multicolumn{3}{c}{Random} & \multicolumn{2}{c}{Ablation} & Ours                                                                                                                                \\ \cmidrule(l){3-5} \cmidrule(lr){6-8} \cmidrule(lr){9-11} \cmidrule(lr){12-13} \cmidrule(lr){14-14}
                                     &                    & Naive                           & ARIMA                        & Holt                       & CNN                          & DeepAR            & CLSTM             & RVFL     & IELM     & SCN               & Stoc-CNN & ES-CNN            & ESM-CNN           \\ \cmidrule(l){1-14}
            \multirow{3}{*}{AR1}     & 1                  & 1.16e+00                        & 2.24e+00                     & 2.23e+00                   & 5.53e-01                     & 5.08e-01          & 4.89e-01          & 3.21e+00 & 5.54e-01 & 3.34e-01          & 4.36e+02 & \textbf{2.28e-01} & \textbf{1.55e-01}\s \\ \cmidrule(l){3-14}
                                     & 3                  & 1.19e+00                        & 2.26e+00                     & 2.25e+00                   & 6.30e-01                     & 7.66e-01          & 6.67e-01          & 5.15e+00 & 5.80e-01 & 4.98e-01          & 4.43e+02 & \textbf{3.22e-01} & \textbf{2.57e-01}\s \\ \cmidrule(l){3-14}
                                     & 6                  & 1.25e+00                        & 2.29e+00                     & 2.28e+00                   & 7.40e-01                     & 9.96e-01          & 6.79e-01          & 5.59e+00 & 6.71e-01 & 6.00e-01          & 1.07e+03 & \textbf{4.61e-01} & \textbf{3.94e-01}\s \\ \cmidrule(l){2-14}
            \multirow{3}{*}{BTC}     & 1                  & 1.47e+04                        & 1.49e+04                     & 1.49e+04                   & 3.87e+03                     & 1.15e+04          & 2.08e+04          & 6.74e+04 & 1.05e+04 & 6.52e+03          & 1.33e+07 & \textbf{2.75e+03} & \textbf{1.59e+03}\s \\ \cmidrule(l){3-14}
                                     & 3                  & 1.47e+04                        & 1.49e+04                     & 1.49e+04                   & 3.97e+03                     & 1.02e+04          & 2.24e+04          & 7.11e+04 & 1.04e+04 & 7.55e+03          & 1.92e+07 & \textbf{2.82e+03} & \textbf{1.74e+03}\s \\ \cmidrule(l){3-14}
                                     & 6                  & 1.46e+04                        & 1.50e+04                     & 1.50e+04                   & 4.00e+03                     & 1.22e+04          & 2.33e+04          & 8.82e+04 & 1.04e+04 & 8.09e+03          & 2.30e+07 & \textbf{2.89e+03} & \textbf{1.92e+03}\s \\ \cmidrule(l){2-14}
            \multirow{3}{*}{ILI}     & 1                  & 9.44e-01                        & 2.23e+00                     & 2.28e+00                   & 8.45e-01                     & 7.82e-01          & 7.10e-01          & 1.74e+00 & 8.88e-01 & 7.14e-01          & 1.50e+01 & \textbf{6.81e-01} & \textbf{6.06e-01}\s \\ \cmidrule(l){3-14}
                                     & 4                  & 1.28e+00                        & 2.25e+00                     & 2.35e+00                   & 1.00e+00                     & 9.95e-01          & 9.07e-01          & 1.96e+00 & 1.04e+00 & 9.09e-01          & 1.39e+01 & \textbf{8.65e-01} & \textbf{8.14e-01}\s \\ \cmidrule(l){3-14}
                                     & 8                  & 1.63e+00                        & 2.36e+00                     & 2.47e+00                   & 1.16e+00                     & 1.16e+00          & \textbf{1.04e+00} & 2.00e+00 & 1.18e+00 & 1.11e+00          & 1.43e+01 & 1.07e+00          & \textbf{1.03e+00}\s \\ \cmidrule(l){2-14}
            \multirowcell{3}{BRENT-\\weekly} & 1                  & 3.24e+01                        & 7.23e+00                     & 7.49e+00                   & 1.01e+01                     & 2.90e+00          & 5.21e+00          & 3.86e+00 & 9.91e+00 & 3.73e+00          & 2.67e+02 & \textbf{2.89e+00} & \textbf{2.62e+00}\s \\ \cmidrule(l){3-14}
                                     & 4                  & 3.21e+01                        & 1.70e+01                     & 1.78e+01                   & 1.12e+01                     & 6.06e+00          & 7.75e+00          & 9.06e+00 & 1.08e+01 & 5.81e+00          & 3.83e+02 & \textbf{5.20e+00} & \textbf{5.14e+00}\s \\ \cmidrule(l){3-14}
                                     & 8                  & 3.31e+01                        & 2.96e+01                     & 3.13e+01                   & 1.12e+01                     & 1.15e+01          & 1.13e+01          & 1.41e+01 & 1.31e+01 & 9.31e+00          & 4.86e+02 & \textbf{7.61e+00} & \textbf{7.52e+00}\s \\ \cmidrule(l){2-14}
            \multirowcell{3}{BRENT-\\daily} & 1                  & 3.14e+01                        & 5.89e+00                     & 5.95e+00                   & 5.22e+00                     & \textbf{1.32e+00}\s & 2.29e+00          & 1.44e+00 & 8.70e+00 & 1.35e+00          & 4.19e+00 & 1.35e+00          & \textbf{1.34e+00} \\ \cmidrule(l){3-14}
                                     & 5                  & 3.17e+01                        & 1.88e+01                     & 1.88e+01                   & 6.19e+00                     & 2.34e+00          & 2.93e+00          & 2.70e+00 & 9.05e+00 & 2.35e+00          & 7.92e+00 & \textbf{2.32e+00} & \textbf{2.29e+00}\s \\ \cmidrule(l){3-14}
                                     & 10                 & 3.21e+01                        & 3.53e+01                     & 3.51e+01                   & 6.26e+00                     & 3.82e+00          & 3.73e+00          & 3.75e+00 & 9.48e+00 & \textbf{3.41e+00} & 1.19e+01 & 3.68e+00          & \textbf{3.16e+00}\s \\ \cmidrule(l){2-14}
            \multirowcell{3}{WTI-\\weekly}   & 1                  & 2.87e+01                        & 7.00e+00                     & 7.31e+00                   & 1.22e+01                     & \textbf{2.64e+00}\s & 5.35e+00          & 4.04e+00 & 1.24e+01 & 3.08e+00          & 2.36e+02 & 3.05e+00          & \textbf{2.81e+00} \\ \cmidrule(l){3-14}
                                     & 4                  & 2.91e+01                        & 1.60e+01                     & 1.68e+01                   & 1.34e+01                     & 5.92e+00          & 8.39e+00          & 9.19e+00 & 1.33e+01 & 5.50e+00          & 3.65e+02 & \textbf{5.21e+00} & \textbf{5.01e+00}\s \\ \cmidrule(l){3-14}
                                     & 8                  & 2.95e+01                        & 2.79e+01                     & 2.94e+01                   & 1.34e+01                     & 8.93e+00          & 1.43e+01          & 1.42e+01 & 1.45e+01 & \textbf{7.51e+00} & 4.25e+02 & 7.65e+00          & \textbf{7.21e+00}\s \\ \cmidrule(l){2-14}
            \multirowcell{3}{WTI-\\daily}   & 1                  & 2.83e+01                        & 5.37e+00                     & 5.43e+00                   & 5.97e+00                     & 1.37e+00          & 2.24e+00          & 1.39e+00 & 8.91e+00 & 1.41e+00          & 6.97e+00 & \textbf{1.37e+00} & \textbf{1.35e+00}\s \\ \cmidrule(l){3-14}
                                     & 5                  & 2.85e+01                        & 1.71e+01                     & 1.69e+01                   & 6.36e+00                     & \textbf{2.24e+00}\s & 2.92e+00          & 2.79e+00 & 9.26e+00 & 2.41e+00          & 1.25e+01 & 2.33e+00          & \textbf{2.32e+00} \\ \cmidrule(l){3-14}
                                     & 10                 & 2.89e+01                        & 3.23e+01                     & 3.15e+01                   & 6.43e+00                     & 3.54e+00          & 3.85e+00          & 4.08e+00 & 9.70e+00 & \textbf{3.39e+00} & 1.74e+01 & 3.84e+00          & \textbf{3.23e+00}\s \\ \cmidrule(l){2-14}
            \multirow{3}{*}{S\&P 500}      & 1                  & 4.24e+02                        & 1.67e+01                     & 2.49e+01                   & 5.48e+01                     & 3.46e+01          & 1.47e+02          & 2.61e+01 & 1.15e+02 & 3.44e+01          & 5.89e+03 & \textbf{1.56e+01} & \textbf{1.28e+01}\s \\ \cmidrule(l){3-14}
                                     & 5                  & 4.20e+02                        & 2.22e+01                     & 3.32e+01                   & 5.77e+01                     & 5.29e+01          & 2.09e+02          & 5.58e+01 & 1.45e+02 & 6.34e+01          & 4.22e+04 & \textbf{2.12e+01} & \textbf{1.93e+01}\s \\ \cmidrule(l){3-14}
                                     & 10                 & 4.23e+02                        & 2.75e+01                     & 4.00e+01                   & 5.49e+01                     & 2.39e+02          & 1.71e+02          & 6.83e+01 & 1.34e+02 & 1.01e+02          & 5.56e+04 & \textbf{2.58e+01} & \textbf{2.43e+01}\s \\ \cmidrule(l){2-14}
            \multirow{3}{*}{NASDAQ}  & 1                  & 1.23e+03                        & 5.42e+01                     & 7.54e+01                   & 2.07e+02                     & 4.99e+01          & 5.17e+02          & 3.20e+02 & 4.30e+02 & 1.29e+02          & 3.70e+05 & \textbf{4.95e+01} & \textbf{4.24e+01}\s \\ \cmidrule(l){3-14}
                                     & 5                  & 1.22e+03                        & 7.41e+01                     & 1.02e+02                   & 2.23e+02                     & 1.57e+02          & 6.80e+02          & 3.44e+02 & 4.51e+02 & 1.79e+02          & 5.97e+05 & \textbf{6.75e+01} & \textbf{6.54e+01}\s \\ \cmidrule(l){3-14}
                                     & 10                 & 1.23e+03                        & 9.38e+01                     & 1.26e+02                   & 2.18e+02                     & 1.19e+03          & 6.56e+02          & 3.19e+02 & 4.41e+02 & 3.16e+02          & 6.22e+05 & \textbf{8.36e+01} & \textbf{8.29e+01}\s \\ \cmidrule(l){2-14}
            \multirow{3}{*}{DJI}     & 1                  & 3.32e+03                        & 1.44e+02                     & 2.09e+02                   & 6.28e+02                     & 2.70e+02          & 1.19e+03          & 4.56e+02 & 1.26e+03 & 2.06e+02          & 2.64e+05 & \textbf{1.30e+02} & \textbf{1.10e+02}\s \\ \cmidrule(l){3-14}
                                     & 5                  & 3.29e+03                        & 2.03e+02                     & 2.87e+02                   & 6.72e+02                     & 1.98e+03          & 1.76e+03          & 6.17e+02 & 1.25e+03 & 5.07e+02          & 2.02e+05 & \textbf{1.84e+02} & \textbf{1.80e+02}\s \\ \cmidrule(l){3-14}
                                     & 10                 & 3.32e+03                        & 2.59e+02                     & 3.58e+02                   & 6.70e+02                     & 1.01e+03          & 1.80e+03          & 1.27e+03 & 1.23e+03 & 1.07e+03          & 2.85e+05 & \textbf{2.36e+02} & \textbf{2.33e+02}\s \\ \cmidrule(l){2-14}
            \multirow{3}{*}{NYSE}    & 1                  & 1.55e+03                        & 8.98e+01                     & 1.33e+02                   & 1.84e+02                     & 1.27e+02          & 2.31e+02          & 9.40e+01 & 3.64e+02 & 1.42e+02          & 1.66e+04 & \textbf{6.97e+01} & \textbf{6.86e+01}\s \\ \cmidrule(l){3-14}
                                     & 5                  & 1.54e+03                        & 1.17e+02                     & 1.76e+02                   & 2.02e+02                     & 2.90e+02          & 3.45e+02          & 2.82e+02 & 3.68e+02 & 1.93e+02          & 2.28e+04 & \textbf{1.05e+02} & \textbf{1.03e+02}\s\\ \cmidrule(l){3-14}
                                     & 10                 & 1.55e+03                        & 1.38e+02                     & 2.10e+02                   & 2.14e+02                     & 1.68e+03          & 4.31e+02          & 4.05e+02 & 3.72e+02 & 2.56e+02          & 3.93e+04 & \textbf{1.28e+02} & \textbf{1.26e+02}\s \\ \bottomrule
        \end{tabular}}
\end{table*}

\end{document}